%% file: main.tex
\documentclass[11pt]{article}
\usepackage[preprint]{acl}

\usepackage{times}
\usepackage{latexsym}
\usepackage[T1]{fontenc}
\usepackage[utf8]{inputenc}
\usepackage{microtype}
\usepackage{inconsolata}
\usepackage{amsmath,amssymb}
\usepackage{booktabs}
\usepackage{array}
\usepackage{tabularx}
\usepackage{graphicx}
\usepackage{placeins}
\usepackage{stfloats}
\usepackage{xcolor}
\usepackage{hyperref}
\usepackage{cleveref}

\newcommand{\E}{\mathbb{E}}
\newcommand{\KL}{D_{\mathrm{KL}}}
\newcommand{\student}{\pi_\theta}
\newcommand{\oldstudent}{\pi_{\theta_{\mathrm{old}}}}

\newcommand{\support}{\Omega_t}
\newcommand{\Vocab}{\mathcal{V}}
\newcommand{\tabrowrule}{\addlinespace[2pt]}

\newcolumntype{L}[1]{>{\raggedright\arraybackslash}p{#1}}
\newcolumntype{Y}{>{\raggedright\arraybackslash}X}
\title{A Formula-Driven Survey and Research Agenda\\
for On-Policy Distillation}

\author{
Bowen Zhang \\
Tsinghua University \\
\texttt{zbw23@mails.tsinghua.edu.cn}
}

\begin{document}
\maketitle

\begin{abstract}
\input{sections/abstract}
\end{abstract}

\input{sections/01_introduction}
\input{sections/02_behavioral_intervention}
\input{sections/03_formula_taxonomy}
\input{sections/04_credit_routing}
\FloatBarrier
\input{sections/05_actionability}
\input{sections/06_failure_mechanisms}
\FloatBarrier
\input{sections/07_diagnostics_recipes}
\FloatBarrier
\input{sections/08_case_studies}
\FloatBarrier
\input{sections/09_open_problems}
\FloatBarrier
\input{sections/10_conclusion}

\FloatBarrier
\bibliography{bib/references}

\clearpage
\appendix
\input{sections/appendix}

\end{document}

%% file: sections/abstract.tex
On-policy distillation (OPD) samples complete or partial trajectories from the current or recent student policy, uses a teacher or self-teacher to score the resulting rollout tokens under their generated contexts, and converts dense log-probability, logit, or distributional signals into post-training updates. This makes OPD a natural interface for modern LLM post-training, but its effectiveness depends on how feedback on student-induced states is made stable and actionable: such feedback can be unreliable, support-truncated, given poor temporal credit across a rollout, used to move probability mass toward weak vocabulary alternatives, or destabilizing. This survey studies OPD as a feedback-to-update problem rather than as a single loss family. We develop a formula-driven taxonomy from two routes, direct distributional losses and policy-gradient-style log-ratio updates, and use it to organize core OPD methods, hybrids with verifiers or outcome signals, industrial reports, framework implementations, failure modes, and stabilization recipes while keeping their evidence levels separate. The taxonomy is also generative. It separates temporal credit assignment from vocabulary-level support and probability-routing choices, which are sometimes bundled together in sampled-token OPD stability discussions. On the temporal-credit axis, we compare immediate, return-to-go, discounted, and baseline-corrected log-ratio estimators, and introduce GAE-OPD as a value-based hypothesis for teacher-student log-ratio returns. On the vocabulary-routing axis, we show why negative sampled-token feedback suppresses a token without specifying a teacher-supported replacement, and introduce Counterfactual Routed OPD (CR-OPD) as a hypothesis for routing probability mass toward teacher-supported, student-reachable alternatives. We close with diagnostics, case-study boundaries, open problems, and a reporting checklist for auditable OPD systems.

%% file: sections/01_introduction.tex
\section{Introduction}
\label{sec:introduction}

Modern LLM post-training mixes supervised imitation, distillation, and reinforcement learning, with feedback ranging from verifiable rewards \citep{deepseekmath2024,deepseekr12025} to rubrics \citep{ropd2026} and teacher signals \citep{qwen32025,mimo2026}. On-policy distillation (OPD) occupies a specific part of this landscape. The current student, or a replay from a recent student policy, first generates a complete or partial trajectory. A teacher or self-teacher then scores tokens along that trajectory, conditioned on the generated history, using dense token-level or distributional feedback. This differs from SFT or teacher-forced KD, where supervision is mostly observed on external or teacher-forced histories. OPD therefore combines the state-coverage motivation of on-policy learning with the information density of distillation.

This interface has growing practical importance. Recent public technical reports describe OPD-style strong-to-weak transfer, multi-teacher expert consolidation, cross-stage anti-regression, and full-vocabulary teacher distillation in several post-training pipelines \citep{qwen32025,mimo2026,glm52026,deepseekv42026}. Public frameworks expose related controls for rollout or replay source, teacher serving, support approximation, divergence choice, rollout buffering, and policy-gradient aggregation \citep{swiftgkd2026,megatronswiftgkd2026,verlopd2026,trlgkdtrainer2026,trldistillationtrainer2026,kdflow2026,tinkerdistillation2026}. Together, these reports and interfaces show that OPD has become implementation-facing post-training infrastructure, where state source, teacher serving, support fidelity, and estimator choice are practical knobs.

The central difficulty is that OPD turns feedback collected along student-induced trajectories into parameter updates. These trajectories reveal the distribution the learner must handle, but they also create an actionability problem: flawed intermediate states, shifted reasoning patterns, tokenizer or support mismatches, and long horizons can make teacher feedback noisy or misaligned \citep{revisitingopd2026,rethinkingopd2026,manyfacesopd2026}; dense local agreement can fail to improve final task success \citep{unmaskingopd2026}; support truncation can hide teacher-supported alternatives \citep{revisitingopd2026,aopd2026}; and sampled-token log-ratio updates can be high variance or suppress a token without specifying a replacement \citep{vopd2026}. OPD is therefore not merely a choice among KL directions or teacher access regimes. It is a feedback-to-update problem in which reliability, support, temporal credit, vocabulary-level routing, and regularization must be specified together.

\paragraph{Why another OPD survey?}
Existing surveys and overviews organize the field by access regime, signal type, divergence family, application domain, or the relationship between off-policy learning, on-policy learning, and reinforcement learning \citep{surveyopd2026,unifiedposttraining2026,opselfdistilloverview2026}. Those views are useful for positioning a fast-moving literature, but they do not directly identify which part of the feedback-to-update path a method changes or which failure it addresses. The same surface label, such as sampled-token, top-$k$, full-vocabulary, black-box, or multi-teacher OPD, can correspond to different state distributions, support choices, credit estimators, gates, update routes, and regularizers. Conversely, methods with different labels may intervene on the same feedback-to-update variable. We therefore start from the objective and estimator formulas, because formulas expose the variables that make OPD methods comparable and composable.

\paragraph{Working definition.}
In this survey, \emph{core OPD} means that student rollouts or recent-policy replay determine the states to be scored, while dense teacher, self-teacher, checkpoint, or support-expanded distributional feedback supplies local token or distributional signals on those states. The resulting update may be a direct local distillation loss, a policy-gradient-style teacher-student log-ratio update, or a hybrid of those two routes. This definition covers GKD-style on-policy distillation \citep{agarwal2024onpolicy}, reverse-KL policy-optimization views \citep{minillm2023}, recent sampled-token or support-expanded OPD analyses \citep{revisitingopd2026}, and the RL-compatible sampled-token framing popularized in the Thinking Machines exposition \citep{thinkingmachines2025}.

\emph{OPD hybrids} keep a dense OPD-style signal but use verifiers, rubrics, reward models, outcome rewards, discriminators, or environment feedback to gate, weight, check, or combine it with more global quality signals \citep{mimo2026,ropd2026,blackboxopd2025,omniopd2026,xu2025kdrl,ssopd2026}. \emph{OPD-adjacent} methods, including pure RLVR, ordinary teacher-forced KD, SFT, or generic verifier training, are outside the core scope unless they explicitly create or modulate dense feedback on student-induced states. Privileged-teacher variants are included when extra context, answers, traces, tools, or visual views are converted into deployable student behavior through dense feedback on student-induced states \citep{opsd2026,opcd2026,visionopd2026}.

\begin{figure*}[!t]
\centering
\includegraphics[width=\textwidth]{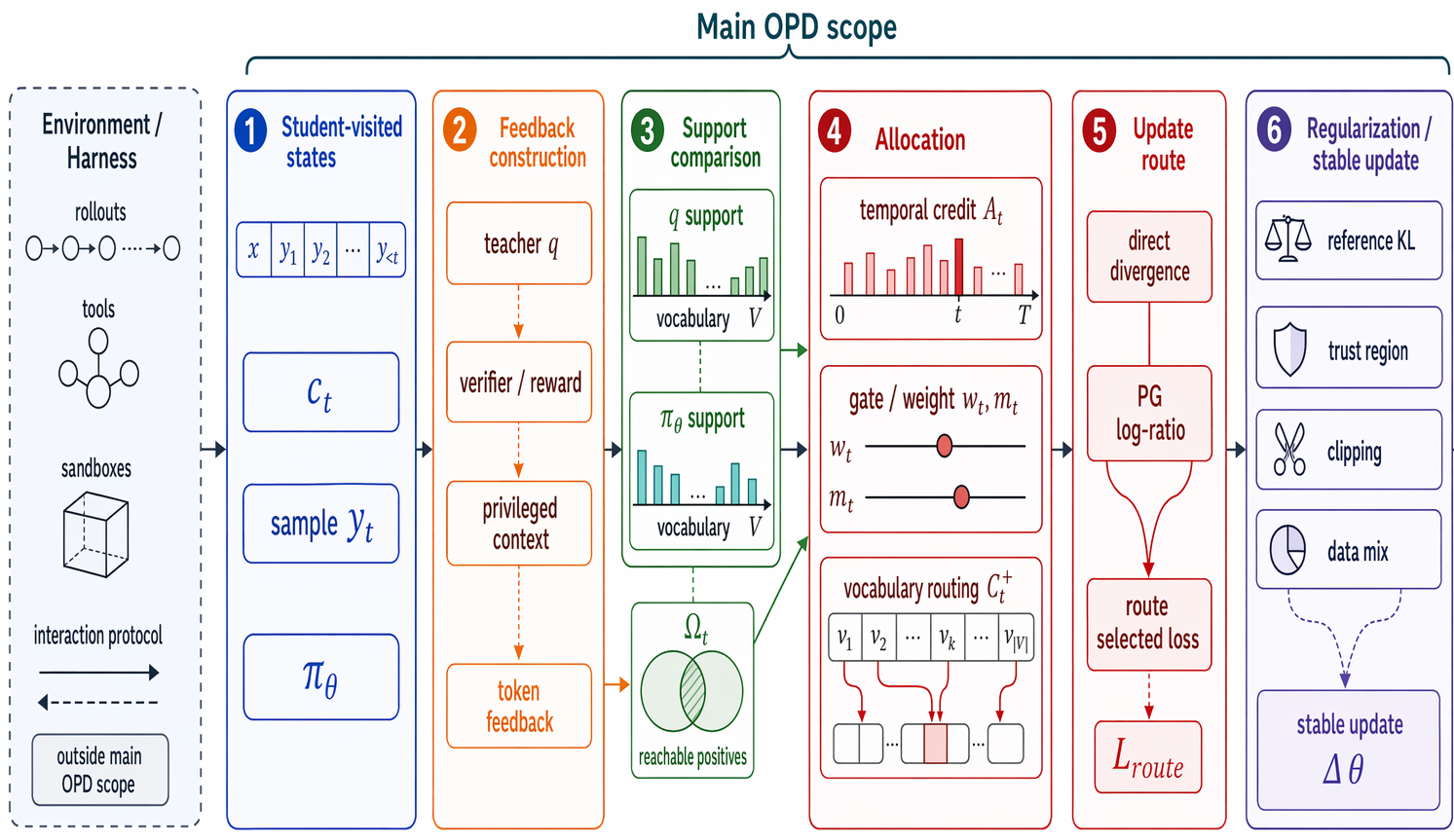}
\caption{Scope boundary of this survey. Environment and harness fidelity are outside the main OPD scope. Given a fixed interface, the survey studies the feedback-to-update path after student rollouts or replay have produced states to score.}
\label{fig:scope-map}
\end{figure*}

This survey also fixes a scope boundary: the environment, evaluator, tool sandbox, and interaction harness are treated as given, and we study what happens after student rollouts or replay have produced states to score. \Cref{fig:scope-map} visualizes this boundary. \Cref{tab:evidence-tiers} states the evidence boundary used throughout the paper. Where compact labels are useful, E0 denotes direct-source papers or reports, E1 official framework documentation, E2 official expositions or blogs, and E3 this survey's synthesis or design hypothesis; \cref{app:fact-audit} gives the detailed audit. Core OPD papers provide evidence about objectives, estimators, support choices, diagnostics, and stabilizers. Industrial and framework reports document public usage and implementation interfaces. Hybrid and adjacent work is used when it changes how dense feedback is gated, weighted, or localized. Multimodal, agentic, online, and cross-tokenizer work is used mainly as scope-expanding evidence unless it provides an explicit dense-feedback mechanism on student-induced states.

\begin{table*}[!t]
\centering
\small
\setlength{\tabcolsep}{5pt}
\renewcommand{\arraystretch}{1.04}
\begin{tabularx}{\textwidth}{L{2.7cm}L{4.0cm}Y}
\toprule
Evidence tier & Representative evidence & How it is used in this survey \\
\midrule
Core OPD evidence & GKD, MiniLLM, Revisiting/Rethinking/Many Faces/Unmasking, AOPD, vOPD, OPD+, TIP, SCOPE, StableOPD, EOPD, OPCD, OPSD. & Grounds objectives, estimators, support choices, diagnostics, and stabilization mechanisms. \\
\tabrowrule
Industrial/system evidence & Qwen3, MiMo, GLM-5, DeepSeek-V4, SWIFT, verl, TRL, KDFlow, Tinker. & Documents public usage, support/cost knobs, teacher-serving constraints, and implementation boundaries. \\
\tabrowrule
Hybrid/adjacent evidence & KDRL, SSOPD, RLSD, SRPO, SD-Zero, Distributional DAgger, verifier/rubric/outcome-routed methods. & Used when sparse or global feedback gates, weights, or localizes dense OPD-style updates. \\
\tabrowrule
Agenda and extension evidence & Cross-tokenizer, multimodal, VLA, agentic, online, cached-teacher, and test-time speculation variants. & Treated as scope-expanding signals and open-problem evidence unless dense feedback on student-induced states is explicit. \\
\bottomrule
\end{tabularx}
\caption{Evidence tiers used to keep core OPD claims separate from system reports, hybrids, adjacent methods, and agenda-setting extensions.}
\label{tab:evidence-tiers}
\end{table*}

\paragraph{Main view.}
We analyze OPD as a feedback-to-update path. First, a method constructs feedback on states induced by student rollouts or replay. Second, it declares the support on which the student and teacher are compared. Third, it allocates that feedback through temporal credit, gates or weights, and vocabulary-level routing. Fourth, it chooses an update route and regularizers that turn the allocated signal into parameter changes. \Cref{sec:taxonomy} derives these variables from direct-loss and policy-gradient formulas, and \cref{fig:update-pipeline} summarizes the resulting map. This formula-driven view is classificatory because it maps existing methods to common design choices, and generative because new OPD variants can be derived by changing or combining variables rather than by naming another global loss.

The clearest generative example appears in \cref{sec:credit-routing}. Sampled-token OPD stability discussions often place temporal credit, support expansion, and estimator baselines in the same design neighborhood: Revisiting OPD couples the sequence/token credit tradeoff with teacher top-$k$ local matching, whereas vOPD separates top-$k$ loss approximation from detached-baseline use \citep{revisitingopd2026,vopd2026}. We separate two mechanisms that enter different parts of the update. Temporal credit asks which scalar should multiply a sampled-token score term: an immediate log-ratio, a return-to-go, a deliberately discounted or windowed surrogate, a state-only baseline, a control variate, or our proposed value-based GAE-OPD advantage. Vocabulary routing asks a different question: after an update decides that local probability should move, which token alternatives should receive the probability mass? Support and vocabulary routing are related but distinct: support determines which alternatives are visible, whereas routing determines where probability mass moves under the selected update. This distinction lets us formulate two taxonomy-derived hypotheses (E3): GAE-OPD for temporal credit assignment and Counterfactual Routed OPD (CR-OPD) for negative-update replacement targets.

The middle of the paper carries these variables into diagnostics. \Cref{sec:actionability} defines feedback actionability as the question of whether a signal should become a gradient on a particular student-induced state. \Cref{sec:failures} reads OPD failures as breakdowns in state compatibility, support reachability, temporal credit, vocabulary routing, privilege compressibility, evaluation alignment, or regularization. \Cref{sec:diagnostics-recipes} maps those failures to conservative stabilization responses using the same variable vocabulary.

\paragraph{Contributions.}
The paper makes the following contributions.
\begin{itemize}
    \item We provide an up-to-date, evidence-tiered survey of OPD for LLM post-training, separating core mechanisms from industrial reports, implementation interfaces, hybrids, adjacent methods, and open-problem evidence.
    \item We introduce a formula-driven taxonomy that derives OPD variables from direct-loss and policy-gradient routes, making method comparisons auditable beyond labels such as sampled-token, top-$k$, full-vocabulary, or multi-teacher OPD.
    \item We use the taxonomy to separate temporal credit from vocabulary-level support and probability routing. This yields bias boundaries for immediate, return-to-go, discounted, and baseline-corrected estimators, and motivates GAE-OPD and CR-OPD as two design hypotheses.
    \item We reinterpret actionability diagnostics, failure mechanisms, stabilization recipes, case studies, and open problems as interventions on the same feedback-to-update variables, ending with a reporting checklist for future OPD work.
\end{itemize}

\paragraph{Roadmap and reading guide.}
\Cref{sec:behavioral-intervention} positions OPD relative to KD, SFT, and RLVR. \Cref{sec:taxonomy} derives the taxonomy; readers mainly interested in method design should read it together with \cref{sec:credit-routing}, which contains the temporal-credit and vocabulary-routing analysis. \Cref{sec:actionability,sec:failures,sec:diagnostics-recipes} form the diagnostic core: when feedback is actionable, how it fails, and which stabilizers target which variables. \Cref{sec:case-studies} is for readers interested in industrial reports and framework knobs under explicit evidence boundaries. \Cref{sec:open-problems} turns the taxonomy into future directions, and \cref{sec:conclusion} summarizes the reporting checklist.

%% file: sections/02_behavioral_intervention.tex
\section{From Distillation to On-Policy Behavioral Intervention}
\label{sec:behavioral-intervention}

Classical post-training already contains two ingredients that OPD recombines, but they should not be collapsed. SFT is local in the sense that every supervised target position can receive a loss, but its target is a one-hot token from the dataset. Ordinary teacher-forced KD is also usually off-policy with respect to the student rollout, yet it is a richer signal than SFT: the teacher distribution can expose alternatives, relative preferences, and uncertainty beyond the observed token \citep{hinton2015distilling}. The shared limitation is not sparsity. It is that these signals are mostly observed on ground-truth, teacher-forced, teacher-generated, or otherwise external histories. During deployment the student conditions on its own previous tokens. This is the classical exposure-bias or state-distribution gap in autoregressive sequence prediction: teacher-forced sequence-to-sequence training provides the canonical baseline against which the mismatch is visible, not itself a direct correction for that gap \citep{sutskever2014sequence}. More direct mismatch responses include scheduled sampling, which mixes model-generated prefixes into training \citep{bengio2015scheduled}, and DAgger-style imitation learning, which queries expert feedback on states visited by the learner \citep{ross2011dagger}. Sequence-level training \citep{ranzato2015sequence} and sequence-level knowledge distillation \citep{kim2016sequence} support the same lesson: dense imitation on expert states can still leave the learner poorly prepared for the states created by its own mistakes.

The on-policy move is therefore a state-coverage intervention. OPD lets the current student, or a recent replay policy, produce complete or partial trajectories, and then scores the generated tokens under the histories that the student actually created. This is different from prefix-masked SFT: the rollout determines the visited states, while the local autoregressive contexts along that rollout are the query points for teacher or self-teacher feedback. The benefit is coverage of self-induced behavior; the cost is that the teacher is now asked to evaluate states that may contain student mistakes, shifted reasoning patterns, or unusual tokenization artifacts.

The distillation move is a feedback-design intervention. A hard SFT label says which token was observed, whereas a soft teacher signal can say which nearby alternatives are plausible, how confident the teacher is, and whether the sampled token is merely suboptimal or outside the teacher's support. OPD inherits this information-density advantage from KD, but using it on student-induced states requires extra choices: which support is visible, which divergence is optimized, whether feedback is used as a direct local loss or as a sampled-token score, how temporal credit is assigned over the trajectory, and whether gates or regularizers should block unreliable feedback.

Reverse-KL OPD is therefore a principled default rather than only a naming convention. The sequence-level objective $\KL(\student(\cdot|x)\|q(\cdot|x))$ places the expectation under the student distribution, so student rollouts are naturally matched to the objective. Its score-function gradient can be written with a detached teacher-student log-ratio reward, which connects reverse-KL distillation to policy-gradient tools and focuses updates on student-supported behavior \citep{minillm2023,thinkingmachines2025}. Forward KL or Jensen-Shannon choices, top-$k$ or full-vocabulary support, and direct-loss versus sampled-token routes answer different support and routing questions: they can preserve teacher-supported alternatives when the support is available, but they are not the same estimator and do not impose the same behavioral bias. Later work makes the choice conditional rather than fixed; for example, EOPD adapts the divergence to token-level distributional properties such as teacher entropy, and AOPD changes the route in non-positive regions where sampled-token reverse-KL feedback is least informative \citep{eopd2026,aopd2026}.

\Cref{tab:regime-boundary} uses this logic to separate state source from feedback-to-update pattern. RLVR and GRPO-style methods \citep{deepseekmath2024,deepseekr12025,dapo2025} already train on student rollouts, but their feedback is often an outcome or group-relative signal rather than a token-local teacher distribution. GLM-5.2 provides recent industrial evidence for the same credit-granularity issue on the neighboring RL side: its official blog reports that long-horizon, compacted traces motivated a move from group-wise optimization to critic-based PPO with token-level advantages rather than group-relative comparisons \citep{zai2026glm52longhorizon}. We treat this as evidence that rollout training can still require finer temporal credit than trajectory-level or group-relative feedback provides. Core OPD occupies the design point where student-induced states receive dense teacher or self-teacher feedback that is converted into local distillation or policy-gradient-style updates. OPD hybrids keep that local distillation update but use verifiers, outcome rewards, rubrics, discriminators, or environment feedback to gate, weight, or combine it with global quality signals \citep{mimo2026,ropd2026,blackboxopd2025,omniopd2026,xu2025kdrl,ssopd2026}.

Recent state-distribution analyses support this separation. \citet{nie2026states} argue that SFT, RL, and OPD differ not only by loss but also by the state distribution on which supervision is applied: fixed data states for SFT, student-induced states with sparse rewards for RL, and student-induced states with richer local feedback for OPD. Distributional DAgger provides an adjacent rich-feedback view in which a learner can query or approximate an expert distribution on states visited by the current policy \citep{agrawal2026distributionaldagger}. These works support our use of states as a taxonomy variable, while leaving open how OPD should weight, route, and regularize the feedback observed on those states.

\begin{table*}[!b]
\centering
\small
\setlength{\tabcolsep}{5pt}
\renewcommand{\arraystretch}{1.08}
\begin{tabularx}{\textwidth}{L{3.05cm}YY}
\toprule
Feedback-to-update pattern & External or teacher-forced states & Student-induced states or rollouts \\
\midrule
Hard token imitation & SFT uses one-hot dataset targets on fixed histories. It is a supervised-imitation baseline, not a soft distillation signal. & DAgger-style queried labels or self-corrections can move hard supervision onto learner states, but this is adjacent unless dense teacher distributions are added. \\
\tabrowrule
Soft distributional distillation & Teacher-forced KD uses teacher probabilities on external histories, preserving more information than hard labels but not correcting the student's rollout distribution. & Core OPD queries teacher, self-teacher, checkpoint, or support-expanded distributions on student rollouts or recent-policy replay. \\
\tabrowrule
Sparse or global outcome signal & Off-policy preference, reward-model, or verifier training can supply global labels, but it is not OPD without dense feedback on visited states. & RLVR and GRPO-style methods train on student rollouts with verifiable, rule-based, or reward-model outcomes; they are adjacent unless they shape dense OPD updates. \\
\tabrowrule
Soft distillation gated by global signal & Usually a data-filtered or verifier-filtered KD baseline. The state source remains mostly external. & OPD hybrids keep the local OPD signal but gate, weight, split, or mix it with verifier, rubric, ORM, discriminator, execution, or outcome feedback. \\
\bottomrule
\end{tabularx}
\caption{Boundary between OPD, neighboring post-training regimes, and OPD hybrids. The relevant axes are the source of states receiving supervision and the pattern by which feedback is allocated and converted into updates.}
\label{tab:regime-boundary}
\end{table*}

Given an input $x$, the student samples a trajectory $y\sim\student(\cdot|x)$ and thereby induces local contexts $c_t=(x,y_{<t})$. We use ``context'' or ``prefix'' here only for the autoregressive conditioning variable at a token position, not for a prefix-masked SFT training unit. In core OPD, the feedback object is usually a next-token distribution, a support-truncated distribution, or a sampled-token log-ratio supplied by a larger teacher, a domain expert checkpoint, a previous-stage checkpoint, or a context-conditioned self-teacher \citep{agarwal2024onpolicy,minillm2023,opcd2026,opsd2026,glm52026,mimo2026}.

Privileged variants differ by what the teacher conditions on, such as extra context, verified traces, answers, tools, or visual views, and by whether that advantage can be compressed into deployable behavior. Verifiers, rubrics, discriminators, and environment-derived outcomes define hybrid OPD signals when they gate, weight, or localize dense feedback on student-induced states \citep{ropd2026,blackboxopd2025,omniopd2026,xu2025kdrl,ssopd2026}.

The historical path to this view passes through several moves. Reverse-KL distillation emphasizes mode-seeking, student-support-focused behavior and connects distillation to policy optimization \citep{minillm2023}; DistiLLM \citep{distillm2024} and DistiLLM-2 \citep{distillm22025} further show that loss geometry and data source should be co-designed. GKD moves distillation onto states induced by self-generated outputs rather than only teacher-forced examples \citep{agarwal2024onpolicy}. Recent OPD work then changes specific variables: sampled-token rewards and top-$k$ support matching \citep{revisitingopd2026}, closed-form control variates \citep{vopd2026}, non-positive-region support matching \citep{aopd2026}, token selection \citep{tip2026}, entropy-aware divergence choice \citep{eopd2026}, and context-conditioned teachers \citep{opcd2026}. These papers motivate the variables formalized next: state source, support, divergence, estimator, gate, routing, and regularizer.

Two axes should therefore be kept separate. One axis asks which process induces the states receiving supervision: teacher-forced data, student rollouts, recent-policy replay, draft-induced states, or environment states. The other asks how feedback is allocated and updated on those states: hard-label imitation, forward KL, reverse KL, Jensen-Shannon divergence, sampled-token log-ratio rewards, support-truncated local matching, outcome-augmented losses, gates, weights, or route switches. A method becomes core or hybrid OPD only after both questions are answered: student-induced states must be paired with dense teacher, self-teacher, checkpoint, or support-expanded feedback, and the route from that feedback to a gradient must be specified. The next section writes these choices as objective variables so that OPD is analyzed as a feedback-to-update interface rather than as a single loss family.

%% file: sections/03_formula_taxonomy.tex
\section{A Formula-Driven Taxonomy of OPD}
\label{sec:taxonomy}

\begin{figure*}[!t]
\centering
\includegraphics[width=\textwidth]{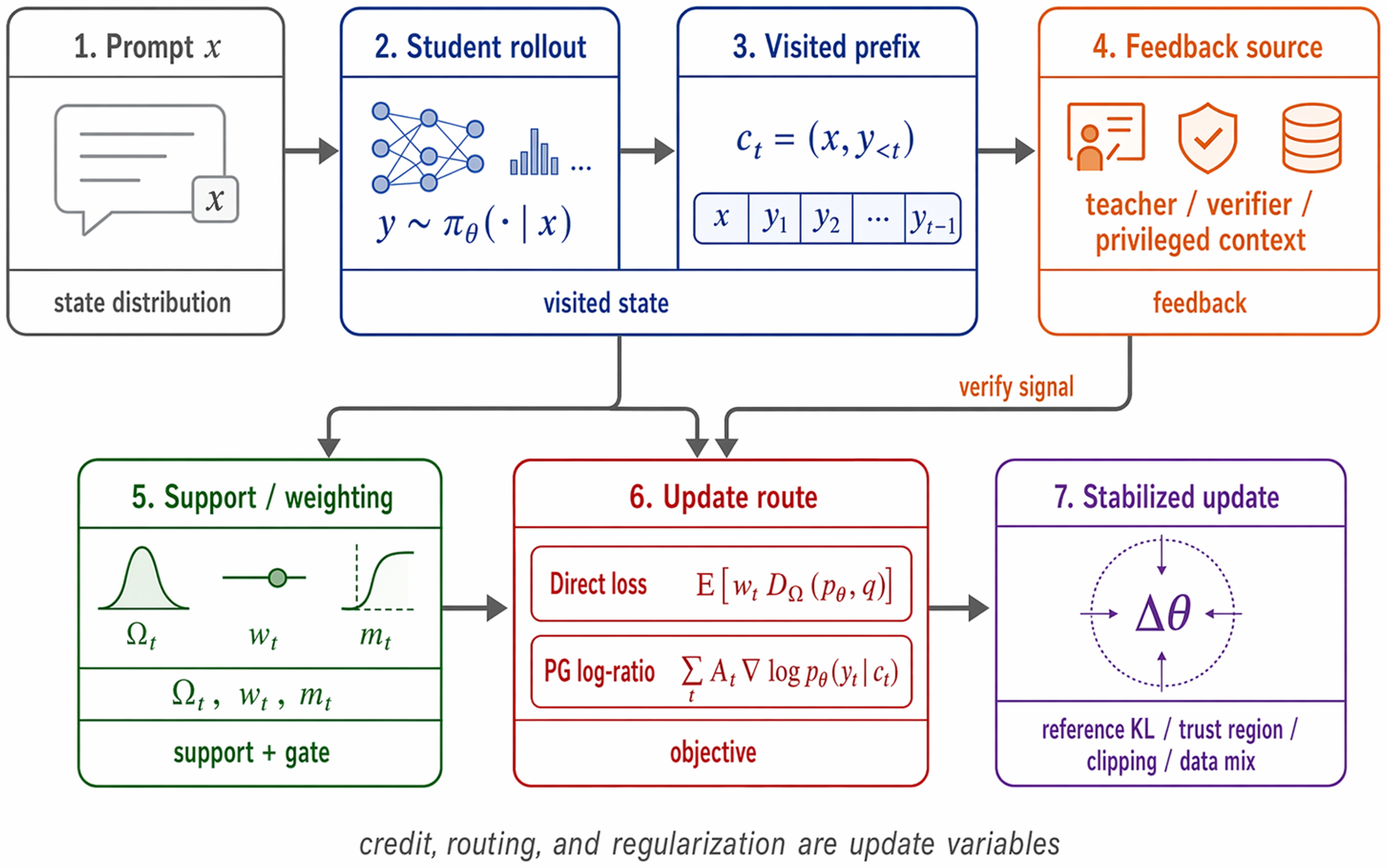}
\caption{Formula-variable map for OPD. The direct-loss route in \cref{eq:direct-opd} and the PG-style route in \cref{eq:token-reward,eq:general-pg} expose the variables summarized in \cref{tab:formula-vars}: state distribution, feedback source, support, temporal credit, gate or weight, vocabulary routing, update route, and regularizer.}
\label{fig:update-pipeline}
\end{figure*}

OPD methods are often described by labels such as sampled-token, top-$k$, full-vocabulary, forward-KL, reverse-KL, white-box, black-box, self-distillation, or multi-teacher. In this survey, we treat these labels as surface descriptions and organize them by variables in the underlying objectives and estimators. We use formulas to expose those variables because they are the handles by which methods respond to OPD failure modes documented in \cref{sec:failures}: unreliable feedback on student-induced states, support mismatch, long-horizon credit variance, directionless negative updates, and drift. \Cref{fig:update-pipeline} gives the pipeline view used throughout this section: each method can be read as choosing a state source, feedback source, support, temporal-credit estimator, gate or weight, vocabulary-routing rule, update route, and regularizer.

We use \emph{feedback-to-update path} as the umbrella term. It has four linked decisions: feedback construction on student-induced states; support construction and comparison, which determine which teacher and student alternatives are observable; allocation through temporal credit and gates or weights; and vocabulary routing plus update and regularization through a chosen gradient route. Support and vocabulary routing are coupled but not identical. Support says which alternatives can be compared; vocabulary routing says where probability mass should move once an update is selected. Within the same path, an \emph{update route} is the differentiation path by which feedback becomes a parameter update, such as a direct local loss, a score-function loss, or a hybrid of the two. A \emph{gate} or \emph{weight} decides whether and how strongly a feedback signal is used, often from correctness, entropy, disagreement, privilege, or verifier information. Temporal credit is the allocation of log-ratio returns across sampled actions. This distinction matters because changing a gate, changing the temporal-credit scalar, changing token alternatives, and changing the differentiation route are different interventions even when all are informally called routing.

\subsection{Direct-Loss Route}

For a prefix $c_t=(x,y_{<t})$, let $\student(v|c_t)$ be the student next-token distribution over vocabulary $\Vocab$, and let $q(v|c_t)$ be a target distribution or feedback-induced distribution. We use the same policy symbol for token and sequence probabilities; the argument identifies the level. Thus $\student(v|c_t)$ is a one-step conditional distribution, while $\student(y|x)$ is the full autoregressive probability of a completion $y=(y_1,\ldots,y_T)$. We write a fixed length for readability; practical implementations either include EOS as an ordinary action or pad to a maximum horizon with masking, so termination and length control remain part of the regularization and state-definition choices:
\begin{equation}
    \student(y|x)
    =
    \prod_{t=1}^{T}
    \student(y_t|x,y_{<t}).
    \label{eq:sequence-factorization}
\end{equation}
When $q$ is an autoregressive teacher or target policy, $q(y|x)=\prod_t q(y_t|x,y_{<t})$ analogously; otherwise $q(\cdot|c_t)$ denotes only the local feedback distribution. A broad class of direct-loss OPD objectives can be written as
\begin{equation}
    \mathcal L_{\mathrm{direct}}
    =
    \E_{c_t\sim d}
    \left[
    w_t
    D_{\support}
    \left[
    \student(\cdot|c_t), q(\cdot|c_t)
    \right]
    \right],
    \label{eq:direct-opd}
\end{equation}
where $d$ is the prefix distribution, $\support$ is the support over which feedback is compared, $D_{\support}$ is a divergence or local matching objective, and $w_t$ is a weight or mask. The argument order in $D_{\support}[\cdot,\cdot]$ is symbolic: the direction of the divergence is itself a design variable. We use $d$ as the rollout or replay distribution from which prefixes are collected; unless stated otherwise, direct-loss updates differentiate the local distributional objective conditional on these prefixes, not the sampling process that produced them. When the prefixes are generated by the current student but treated as fixed training data, direct-loss OPD is a semi-gradient with respect to the full on-policy objective.

\paragraph{Degrees of freedom.}
In summary, direct-loss OPD names the differentiation route rather than the KL direction. \Cref{eq:direct-opd} exposes the prefix law $d$; the local target $q$ and any teacher privilege used to construct it; the comparison support $\support$ and its normalization; the local objective $D_{\support}$, including argument order and KL direction; the mask or weight $w_t$; and the gradient path, namely which of $d$, $q$, $\support$, and $w_t$ are treated as fixed rollout data versus differentiated quantities. Regularizers such as reference KL, entropy control, clipping, or golden-data mixing are separate outer choices that may be added to this local matching term.

\subsection{Policy-Gradient Route}

The policy-gradient route starts from the sequence-level reverse-KL objective and uses the fact that its gradient has a score-function form. Among the works covered in this survey, MiniLLM \citep{minillm2023} is a central early LLM-distillation precursor for this route.

It casts sequence-level reverse-KL distillation as a policy-gradient estimator and uses stabilization for the resulting high-variance update. For a single input, the sequence objective is
\begin{equation}
\begin{aligned}
    J_x(\theta)
    &=
    \KL(\student(\cdot|x)\|q(\cdot|x))
    \\
    &=
    \E_{y\sim\student}
    \left[
    \log \student(y|x)
    -
    \log q(y|x)
    \right].
\end{aligned}
    \label{eq:sequence-rkl}
\end{equation}
Because the sampling distribution depends on $\theta$, differentiating \cref{eq:sequence-rkl} is not the same as backpropagating through a fixed supervised loss. Applying the score-function identity yields
\begin{equation}
\begin{aligned}
    \nabla_\theta J_x
    &=
    \E_{y\sim\student}
    \Big[
    \big(
    \log \student(y|x)-\log q(y|x)+1
    \big)
    \\
    &\qquad\qquad\cdot
    \nabla_\theta \log \student(y|x)
    \Big].
\end{aligned}
    \label{eq:rkl-gradient}
\end{equation}
The additional $+1$ term is a zero-mean constant score term because the expected score is zero. Equivalently, the reverse-KL update can be written with reward $R(y)=\log q(y|x)-\log \student(y|x)$ after dropping this constant-score control-variate term. Dropping it preserves the expected gradient under exact on-policy sampling, but it can change finite-sample variance. In practical PG-style OPD losses, this reward or advantage is detached before multiplying the student log-probability; otherwise the implementation silently changes the differentiated objective.

At token granularity, sampled-token implementations often use the immediate teacher-student log-ratio
\begin{equation}
    r_t
    =
    \log q(y_t|c_t)
    -
    \log \student(y_t|c_t),
    \label{eq:token-reward}
\end{equation}
for a sampled token $y_t\sim \student(\cdot|c_t)$. Positive $r_t$ means that the teacher assigns higher probability to the sampled token than the student does. This log-ratio assumes that the teacher assigns positive mass on the sampled or declared support. If teacher probabilities can be zero, hidden by an API, or truncated away, the implementation must state its smoothing, clipping, or support convention before $r_t$ can be interpreted as a finite reward.

\paragraph{Fixed-prefix local objective.}
It is useful to separate a token-local distributional objective from the sequence objective above. With fixed prefix $c_t$ and detached teacher distribution $q(\cdot|c_t)$, full-vocabulary reverse KL is finite only if $q(v|c_t)>0$ whenever $\student(v|c_t)>0$ on the active vocabulary. Under this absolute-continuity convention, the local objective is
\begin{equation}
\begin{aligned}
    D_t^{\mathrm{RKL}}
    &=
    \KL(\student(\cdot|c_t)\|q(\cdot|c_t))
    \\
    &=
    \sum_{v\in\Vocab}
    \student(v|c_t)
    \log
    \frac{\student(v|c_t)}{q(v|c_t)} .
\end{aligned}
    \label{eq:local-rkl}
\end{equation}
This is the finite-KL condition, not an extra modeling preference. It usually holds for full-logit softmax teachers on a shared unmasked vocabulary. If it fails, \cref{eq:local-rkl} is no longer a finite full-vocabulary KL: smoothing keeps the same form with a smoothed target, while truncation changes the object to a normalized restricted-support divergence $D_{\Omega_t}$.

Writing $\E_t$ for expectation over $a\sim\student(\cdot|c_t)$, use the shorthand
\[
\begin{aligned}
    g_t(a)
    &=
    \nabla_\theta\log\student(a|c_t),
    \\
    \rho_t(a)
    &=
    \log q(a|c_t)-\log \student(a|c_t).
\end{aligned}
\]
Under the same conventions, the score identity is
\begin{equation}
\begin{aligned}
    -\nabla_\theta D_t^{\mathrm{RKL}}
    &=
    \E_t
    \left[
    \left(\rho_t(a)-1\right)
    g_t(a)
    \right]
    \\
    &=
    \E_t
    \left[
    \rho_t(a)
    g_t(a)
    \right],
    \\
    \E_t[g_t(a)]
    &=
    0 .
\end{aligned}
    \label{eq:local-rkl-score}
\end{equation}
The second equality uses the zero-mean score identity in the last line, so the constant $-1$ term has zero expectation. Thus a sampled-token immediate update is an unbiased one-sample estimator of the fixed-prefix local reverse-KL descent direction, while it is generally a biased surrogate for the sequence-level reverse-KL direction because it omits future log-ratio terms. Full-vocabulary local OPD computes \cref{eq:local-rkl} with $\Omega_t=\Vocab$ and $q(v|c_t)>0$ for all student-supported $v$. Support-truncated variants replace $\Vocab$ by a declared $\Omega_t$ and must state smoothing and whether $q$, $\student$, or both are renormalized. Sampled-token OPD estimates \cref{eq:local-rkl-score}; sequence-level PG-style OPD uses return-to-go credit over the whole trajectory.

The score-function form is the reason PG-style OPD naturally inherits the RL estimator stack. Writing $\E_t$ for expectation over $y_t\sim \student(\cdot|c_t)$, any state-only baseline $b(c_t)$ satisfies
\begin{equation}
\begin{aligned}
    \E_t
    [
    b(c_t)
    \nabla_\theta \log \student(y_t|c_t)
    ]
    &=
    b(c_t)
    \sum_{y_t}
    \nabla_\theta \student(y_t|c_t)
    \\
    &=
    b(c_t)
    \nabla_\theta
    \sum_{y_t}
    \student(y_t|c_t)
    \\
    &=
    0,
\end{aligned}
    \label{eq:baseline-identity}
\end{equation}
so log-ratio returns can be replaced by baseline-subtracted advantages without changing the exact conditional on-policy expectation. This expectation-preserving property is specific to state-only additive baselines. Multiplying future log-ratios by a discount, truncating the horizon, or using bootstrapped residuals with an approximate critic changes the estimator unless the corresponding objective and value-function assumptions are stated. Baselines, control variates, PPO-style clipping \citep{schulman2017ppo}, and trust regions \citep{schulman2015trpo} are therefore not cosmetic additions; they are the natural variance-control and stability interface once reverse-KL distillation is estimated as a policy gradient.

A general sampled-token OPD parameter-update direction can be written as
\begin{equation}
    \Delta\theta_{\mathrm{OPD}}
    \propto
    \sum_t
    A_t
    \nabla_\theta \log \student(y_t|c_t),
    \label{eq:general-pg}
\end{equation}
where the expression corresponds to descending reverse KL when $A_t$ estimates teacher-student log-ratio reward. $A_t$ may be an immediate log-ratio reward, a return-to-go, a deliberately discounted or windowed surrogate, a baseline-corrected advantage, or a control-variate estimator; we also analyze a value-based GAE-style design point \citep{schulman2016gae} in \cref{sec:credit-routing}. These choices and design points differ in bias conditions: immediate, truncated, or discounted returns are generally biased relative to the undiscounted sequence-level reverse-KL direction, whereas state-only baselines preserve the exact on-policy expectation; the proposed value-based GAE-style design point would preserve that direction under exact on-policy values and detached rewards, while $\lambda=1$ has the special terminal Monte Carlo property of telescoping to a return minus a state-only baseline even when the baseline value is approximate.
If implemented as a loss minimized by gradient descent, the corresponding policy term has the opposite sign,
\begin{equation}
    \mathcal L_{\mathrm{PG}}
    =
    -
    \sum_t
    \operatorname{sg}(A_t)
    \log \student(y_t|c_t),
\end{equation}
where $\operatorname{sg}(\cdot)$ denotes stop-gradient.

\paragraph{Degrees of freedom.}
In summary, PG-style OPD exposes the rollout policy used to collect actions, such as current, cached, replayed, or importance-corrected policies; the reward definition, such as sequence log-ratio, immediate token log-ratio, clipped or normalized variants, or mixtures with outcome rewards; the credit estimator, such as immediate, return-to-go, discounted or windowed, value-baseline, control-variate, or proposed GAE-style advantages; the stop-gradient and old-policy treatment of rewards; the support used when feedback goes beyond the sampled token; and stabilizers such as PPO-style clipping, trust regions, reference KL, entropy control, and length control. These choices are independent of the surface label ``sampled-token'': two sampled-token methods can differ mainly in baseline, credit horizon, reward clipping, or route switching.

\subsection{Variables Exposed by the Formulas}

\begin{table*}[!b]
\centering
\footnotesize
\setlength{\tabcolsep}{4pt}
\setlength{\abovecaptionskip}{4pt}
\renewcommand{\arraystretch}{1.0}
\begin{tabularx}{\textwidth}{L{2.15cm}L{2.7cm}L{5.2cm}Y}
\toprule
Variable & Question & Observed choices and design points & Main risk \\
\midrule
State / rollout source & Which process induces the states being scored? & Teacher-forced states, student rollouts, mixed data/rollout states, recent-policy replay, or environment states. & Exposure gap or flawed-state trap. \\
\tabrowrule
Feedback source & Who provides local targets? & Teacher logits, self-teacher logits, previous checkpoint, domain expert, support-truncated target, or a global signal used to gate dense feedback. & Fluent but wrong local feedback. \\
\tabrowrule
Teacher privilege & What does the feedback source know? & Larger model, previous checkpoint, domain expert, answer, context, tool trace, or visual crop. & Uncompressible privilege or leakage. \\
\tabrowrule
Support & Where is feedback compared? & Sampled token, teacher top-$k$, student top-$k$, overlap, full vocabulary, aligned text/span support, hidden-state support, or logit-free surrogate. & Truncation, mismatch, or missing alternatives. \\
\tabrowrule
Update route & How does feedback become gradient? & Direct KL, PG log-ratio, hybrid KL+PG, sign-conditioned switch, or global-feedback-gated dense update. & High variance or objective-support mismatch. \\
\tabrowrule
Temporal credit estimator & How far do log-ratio returns propagate? & Immediate token, return-to-go, discounted or windowed return, baseline, control variate, or proposed GAE-style advantage \citep{schulman2016gae}. & Bias-variance failure, critic error, or off-policy drift. \\
\tabrowrule
Gate or weight & Which feedback is used, and how strongly? & Entropy/disagreement weights, correctness weights, verifier gates, privilege gates, trajectory filters, or token masks. & Overweighted noise or discarded useful signal. \\
\tabrowrule
Vocabulary probability routing & Where does local probability mass move? & Implicit softmax, local divergence switch, teacher top-$k$, full vocabulary, overlap support, or routed alternatives. & Negative feedback without a replacement target. \\
\tabrowrule
Regularizer & How is drift controlled? & Reference KL, golden-data mixture, clipping, trust region, entropy control, or length control. & Repetition, length inflation, or forgetting. \\
\bottomrule
\end{tabularx}
\caption{A formula-derived taxonomy of OPD variables. The entries are design dimensions whose useful settings depend on the state, feedback source, support, and update route.}
\label{tab:formula-vars}
\end{table*}

The variables in \Cref{tab:formula-vars} separate optimization routes from divergence choices. They also subsume several recurring second-order handles. Outcome or verifier coupling is a choice of feedback source, weight, and update route rather than a separate OPD axis \citep{mimo2026,xu2025kdrl,ssopd2026,rwopd2026}. Teacher assignment and teacher freshness change the feedback distribution $q$, its mixture weights, and the rollout or cache schedule \citep{deepseekv42026,madopd2026,lightningopd2026,nearpolicy2026}. Granularity chooses whether $w_t$, $A_t$, and support masks act at token, span, step, trajectory, or skill level \citep{fireopd2026,sodagents2026,gear2026}. Calibration and uncertainty objectives are regularization or evaluation-alignment choices distinct from better local imitation \citep{certaintyopd2026,egrsd2026}. We keep these handles nested under the main variables so that the taxonomy remains compact while still explaining multi-teacher, verifier-guided, agentic, and online variants.

Direct-loss OPD differentiates a local distributional objective on a chosen support. PG-style OPD treats teacher-student log-ratios as rewards on sampled actions. The two routes can be combined: for example, methods may use policy-gradient updates for positive-advantage samples and switch to teacher-support matching in non-positive regions \citep{aopd2026}. \Cref{tab:route-comparison} summarizes the operational difference and the minimum variables that make a reported OPD loop auditable.

\paragraph{Implementation assumptions.}
Several implementation choices change the mathematical object being optimized. First, $q(\cdot|c_t)$ is treated as detached feedback unless a method explicitly trains a co-evolving teacher; external teachers, previous checkpoints, self-teachers, and privileged teachers may be constructed differently, but the student loss should state whether teacher logits, masks, and weights are stop-gradient. Second, direct-loss OPD on student rollouts is usually a local semi-gradient: the collected prefixes are fixed data for the update, even if a current or recent student generated them. Third, support truncation requires an explicit normalization convention. Forward KL on teacher top-$k$ support, for example, may renormalize only the teacher target, or renormalize both teacher and student on the restricted support; these choices differ from full-softmax KL because omitted vocabulary mass receives different gradients. Fourth, PG-style OPD must distinguish rewards computed from the current student from cached rewards computed under an old policy. Using current log-ratios, stale cached log-ratios, or importance-corrected log-ratios defines different estimators and different bias terms. Concrete instantiations include teacher-top-$k$ forward KL as a direct loss, full-vocabulary or support-truncated reverse KL as a direct loss, sampled-token log-ratio reward as a PG estimator, and AOPD-style sign-conditioned switching between positive-advantage PG and non-positive-region local divergence matching \citep{aopd2026}.

\FloatBarrier

The support variable is where these mathematical choices meet systems constraints. Higher-fidelity or better-aligned support usually spends more on logit materialization, teacher serving, transfer, alignment, or routing. The main text keeps the distinction conceptual: sampled-token support is cheap but blind to alternatives; top-$k$ and overlap supports restore local alternatives with truncation and normalization choices; full-vocabulary support is most faithful but expensive; cross-tokenizer and representation supports require an explicit event bridge; and logit-free feedback replaces raw vocabulary support with a localization route. \Cref{tab:support-cost} gives the detailed support ladder. \Cref{eq:cross-tokenizer-bridge} gives one abstract cross-tokenizer bridge, in the appendix.

\subsection{Representative Methods under the Taxonomy}

This subsection highlights the variables changed by representative methods. Foundational work sets the state, objective, and route variables: GKD moves distillation from fixed teacher-forced states toward states induced by student-generated outputs, including mistakes \citep{agarwal2024onpolicy}; MiniLLM connects reverse-KL distillation to policy-gradient estimators \citep{minillm2023}; DistiLLM \citep{distillm2024} and DistiLLM-2 \citep{distillm22025} show that loss geometry, data source, and contrastive structure should be co-designed. Recent state-distribution \citep{nie2026states} and rich-feedback analyses \citep{agrawal2026distributionaldagger} make the same point more explicitly: post-training methods differ in where supervision is applied, not only in which loss is minimized.

\Cref{tab:method-taxonomy} gives a compact method map in the main text, while the detailed mechanism map in \cref{tab:method-taxonomy-detailed} and the expanded literature matrices in \cref{app:literature-matrix} keep extension families, industrial systems, framework implementations, and implementation-facing details auditable. The lesson of the compact map is that method labels are often misleading. vOPD \citep{vopd2026} and AOPD \citep{aopd2026} are both ``sampled-token'' variants in a surface sense, but they intervene on different variables: vOPD changes the estimator baseline, while AOPD changes the route for non-positive regions. TIP \citep{tip2026} and SCOPE \citep{scope2026} are both weighting methods, but TIP selects informative token positions, whereas SCOPE routes trajectories by correctness. OPCD \citep{opcd2026} and Vision-OPD \citep{visionopd2026} are both privileged-teacher methods, but the relevant risk is not privilege itself; it is whether the privileged signal can be compressed into behavior available at deployment.

\begin{table}[!htbp]
\centering
\scriptsize
\setlength{\tabcolsep}{3.5pt}
\setlength{\abovecaptionskip}{4pt}
\renewcommand{\arraystretch}{1.0}
\begin{tabularx}{\linewidth}{L{2.45cm}Y}
\toprule
Mechanism family & Variables and actionability lesson \\
\midrule
Foundations and diagnostics \citep{agarwal2024onpolicy,minillm2023,revisitingopd2026} & Change the state source, divergence, route, or feedback-reliability test; OPD is not only a new loss, but supervision on student-induced states. \\
\tabrowrule
Estimator and temporal-credit controls \citep{vopd2026,opdplus2026} & Change baselines, advantage construction, stop-gradient treatment, or horizon; the main tradeoff is bias--variance under teacher reliability. \\
\tabrowrule
Route switching and replacement routing \citep{aopd2026,tgpo2026} & Change the update route or vocabulary target for non-positive feedback; suppressing a sampled token is different from naming its replacement. \\
\tabrowrule
Gates, weights, and adaptive objectives \citep{tip2026,scope2026,stableopd2026} & Change where or how strongly feedback is used; actionability may depend on positions, trajectories, entropy, disagreement, or correctness gates. \\
\tabrowrule
Outcome, privilege, and logit-free feedback \citep{blackboxopd2025,opcd2026,visionopd2026} & Change the feedback source or teacher privilege; global or privileged signals must be localized and compressible into deployment behavior. \\
\tabrowrule
Support and systems \citep{deepseekv42026,swiftgkd2026,verlopd2026} & Change support fidelity, cache freshness, teacher serving, or framework knobs; systems choices are part of the objective interface. \\
\bottomrule
\end{tabularx}
\caption{Compact method matrix under the formula-driven taxonomy. Rows group representative mechanisms by the variables they primarily change; expanded matrices and claim audits remain in the appendix.}
\label{tab:method-taxonomy}
\end{table}

\subsection{Evidence Discipline}

Throughout this survey, we distinguish three kinds of statements. A literature-supported claim is one that prior work explicitly reports; we attribute it. A synthesis is our organization of several findings under the taxonomy above; we mark it as interpretation. A design direction is a potential OPD variant derived from the taxonomy; we state its assumptions and evaluation requirements. This distinction matters because many OPD diagnostics, such as top-$k$ overlap or teacher recoverability on student-induced states, are useful correlates or predictors whose interpretation depends on the teacher, student, task, and support regime.

\begin{table*}[!t]
\centering
\footnotesize
\setlength{\tabcolsep}{4pt}
\setlength{\abovecaptionskip}{4pt}
\renewcommand{\arraystretch}{1.0}
\begin{tabularx}{\textwidth}{L{2.8cm}L{3.2cm}L{3.2cm}Y}
\toprule
Route & Support and gradient object & Credit or gate & Must report \\
\midrule
Direct-loss route & Full vocabulary, teacher top-$k$, student top-$k$, overlap, or aligned support; differentiate a local distributional objective conditional on collected prefixes. & Usually local at a prefix unless the target distribution or weight encodes future or global information. & Support source and size, normalization convention, divergence direction, masks or weights, detached prefixes/targets, and drift regularizer. \\
\tabrowrule
PG-style route & Sampled token by default; top-$k$ or routed alternatives require an explicit extra support query. Differentiate a score-function loss over sampled actions. & Immediate, return-to-go, discounted, windowed, baseline-corrected, control-variate, or proposed GAE-style advantages \citep{schulman2016gae}. & Rollout policy, teacher freshness, reward definition, horizon, baseline/control variate, stop-gradient convention, clipping or trust region, and negative-advantage routing. \\
\tabrowrule
Hybrid or gated route & Dense teacher/self-teacher feedback plus verifier, rubric, ORM, discriminator, execution, or environment signal. The route may switch by sign, correctness, entropy, or support overlap. & The global signal gates, weights, splits, or localizes dense updates; correct and incorrect rollouts may follow different routes. & Which signal is dense, which is global, how the global signal is localized, whether the method remains core OPD or hybrid, and whether gains are task gains or teacher-agreement gains. \\
\bottomrule
\end{tabularx}
\caption{Auditable OPD update routes. The distinction is about how the objective is estimated, differentiated, and gated, not about a fixed KL direction.}
\label{tab:route-comparison}
\end{table*}

%% file: sections/04_credit_routing.tex
\section{Temporal Credit and Vocabulary Probability Routing}
\label{sec:credit-routing}

The taxonomy separates two variables that are sometimes discussed together under sampled-token OPD stability, but that enter different parts of the update. Here \emph{routing} means vocabulary-level probability routing, not the broader update route or global gate defined in \cref{sec:taxonomy}. The first variable is \emph{temporal credit}: which scalar should multiply the score term $s_t=\nabla_\theta \log \student(y_t|c_t)$ for a sampled token? The second variable is \emph{vocabulary routing}: once an update says that probability should move at prefix $c_t$, which token alternatives should receive that probability mass? 

\begin{figure*}[t]
\centering
\includegraphics[width=0.95\textwidth]{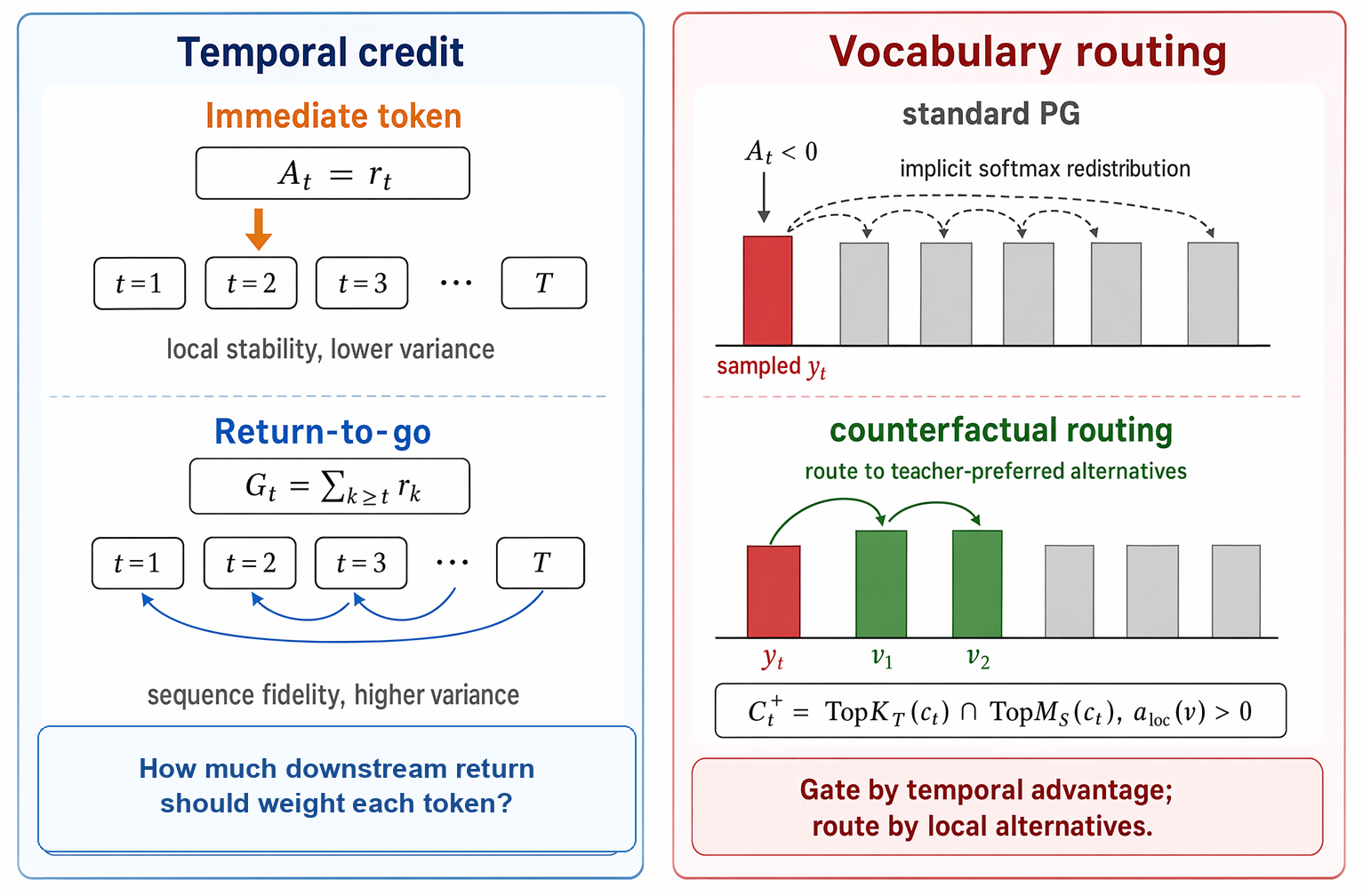}
\caption{Two update axes exposed by PG-style OPD. Temporal credit asks how downstream log-ratio returns should weight each sampled action; vocabulary routing asks where probability mass should move after a negative sampled-token update. The counterfactual-routing panel illustrates the CR-OPD hypothesis.}
\label{fig:credit-routing}
\end{figure*}

Return-to-go, discounting, and state-only baselines are temporal-credit estimator moves; teacher top-$k$, full-vocabulary KL, and AOPD-style switching address vocabulary routing. This section then analyzes two potential design directions in parallel: GAE-OPD, an OPD adaptation of generalized advantage estimation \citep{schulman2016gae}, for value-based temporal credit, and Counterfactual Routed OPD (CR-OPD) for vocabulary-level replacement targets. The separation is mathematical: these mechanisms enter different parts of the PG-style update. \Cref{fig:credit-routing} summarizes this distinction.

Recent OPD work motivates this separation. Revisiting OPD analyzes the bias--variance trade-off between sequence-level and token-level estimators, then motivates teacher top-$k$ local support matching as a less brittle local signal \citep{revisitingopd2026}. That top-$k$ objective is a local support-truncated surrogate, while vOPD makes a complementary estimator distinction: a top-$k$ KL approximation changes the objective when used as a loss, but can preserve the conditional score expectation when used only as a detached action-independent baseline \citep{vopd2026}. The return-to-go algebra, token-local bias term, discount-versus-baseline distinction, and logit-level routing analysis below are formula consequences under the stated assumptions. Teacher top-$k$ matching, AOPD-style switching, vOPD-style control variates, and related stabilizers are literature-supported mechanisms. GAE-OPD and CR-OPD are taxonomy-derived hypotheses (E3) whose assumptions and evaluation requirements follow from those same formulas.

\subsection{From Sequence Reverse KL to Return-to-Go}

We use the reward convention from Eq.~\ref{eq:token-reward},
\begin{equation}
    r_t
    =
    \log q(y_t|c_t)-\log \student(y_t|c_t),
\end{equation}
so ascending reward corresponds to descending the reverse-KL cost up to the zero-mean constant score term in Eq.~\ref{eq:rkl-gradient}. When $q$ supplies autoregressive probabilities on the declared support, the completion-level reward for $y=(y_1,\ldots,y_T)$ is
\begin{equation}
    R(y)
    =
    \sum_{t=1}^{T} r_t
    =
    \log q(y|x)-\log \student(y|x),
\end{equation}
and autoregressive factorization gives
\begin{equation}
    \nabla_\theta \log \student(y|x)
    =
    \sum_{t=1}^{T}s_t .
\end{equation}
The naive sequence estimator multiplies every score term by the full sequence reward. Past rewards can be removed from the score term at time $t$ without changing the expectation: once $c_t$ is fixed, $r_1,\ldots,r_{t-1}$ are fixed with respect to the sampled action $y_t$, and
\begin{equation}
    \E_{y_t\sim \student(\cdot|c_t)}[s_t]=0.
\end{equation}
Thus, under exact current-policy sampling, compatible teacher support, and detached log-ratio weights, the sequence-level reverse-KL descent direction can be estimated by the return-to-go form
\begin{equation}
    \hat g_{\mathrm{seq}}
    =
    \sum_{t=1}^{T}
    G_t s_t,
    \qquad
    G_t=\sum_{k=t}^{T} r_k .
    \label{eq:return-to-go}
\end{equation}
This is ordinary causal credit assignment: the action at time $t$ can influence suffix states and therefore downstream log-ratios, while past tokens remain fixed under the current prefix. If rollouts are generated by an old policy and reused for multiple updates, Eq.~\ref{eq:return-to-go} becomes off-policy unless importance correction, PPO-style clipping \citep{schulman2017ppo}, or a trust-region approximation \citep{schulman2015trpo} is added.

The problem is that OPD's reward is not an environment truth signal. It is teacher-student log-ratio feedback evaluated on student-induced states. In long reasoning or agentic trajectories, later states are increasingly shaped by the student; teacher reliability on such states is a state-compatibility risk highlighted by recent OPD diagnostics \citep{revisitingopd2026,rethinkingopd2026,manyfacesopd2026}. Downstream log-ratios therefore create two coupled risks: they increase estimator variance, and they may become less reliable as teacher feedback. Revisiting OPD analyzes the variance side: under bounded per-token log-ratio and score terms, token-local sums have a much tighter worst-case horizon dependence than sequence-level estimators, with simplified bounds of $O(T^2)$ versus $O(T^4)$ \citep{revisitingopd2026}.

\subsection{Token-Level and Discounted Estimators Are Biased Surrogates}

Immediate sampled-token OPD uses
\begin{equation}
    \hat g_{\mathrm{tok}}
    =
    \sum_{t=1}^{T}
    r_t s_t .
    \label{eq:token-estimator}
\end{equation}
At a fixed prefix, the same sampled term is an unbiased one-sample estimator of the local reverse-KL direction in \cref{eq:local-rkl-score}. The bias discussed here is relative to the sequence-level objective: compared with Eq.~\ref{eq:return-to-go}, it drops all future terms. The dropped term is
\begin{equation}
    \hat g_{\mathrm{seq}}-\hat g_{\mathrm{tok}}
    =
    \sum_{t=1}^{T}
    \sum_{k=t+1}^{T}
    r_k s_t .
    \label{eq:token-bias-term}
\end{equation}
Its expectation is generally not zero because $r_k$ depends on the suffix distribution induced by the earlier sampled action $y_t$. Token-level OPD is therefore a biased low-variance local surrogate for the sequence-level reverse-KL direction, not an unbiased estimator of that direction.

A discounted return estimator makes the same tradeoff continuously:
\begin{equation}
\begin{aligned}
    \hat g_{\gamma}
    &=
    \sum_t
    \left(
    \sum_{k=t}^{T}
    \gamma^{k-t}r_k
    \right)s_t,
    \quad
    \gamma\in[0,1].
\end{aligned}
    \label{eq:discounted}
\end{equation}
$\gamma=0$ gives Eq.~\ref{eq:token-estimator}; $\gamma=1$ gives Eq.~\ref{eq:return-to-go}. For $\gamma<1$, however, the estimator changes the expected direction:
\begin{equation}
\begin{aligned}
    \E[\hat g_{\mathrm{seq}}-\hat g_{\gamma}]
    =
    \E
    \left[
    \sum_{t=1}^{T}
    \sum_{k=t+1}^{T}
    (1-\gamma^{k-t})r_k s_t
    \right],
\end{aligned}
    \label{eq:discount-bias}
\end{equation}
which is generally nonzero. This is the key distinction: a state-only baseline is subtracted and has zero expected score contribution; a discount factor multiplies future rewards and changes the objective. Discounting can still be sensible in OPD, but it should be named correctly. It is a biased temporal-credit surrogate, or a reliability discount when later teacher feedback on student-induced states is suspected to be noisier or less actionable.

\subsection{State-Only Baselines and a Proposed GAE-Style Extension}

Baselines have a different mathematical status from discounting. For any state-only function $b(c_t)$,
\begin{equation}
    \E_{y_t\sim\student(\cdot|c_t)}
    [
    b(c_t)s_t
    ]
    =
    b(c_t)
    \nabla_\theta
    \sum_{y_t}\student(y_t|c_t)
    =
    0.
\end{equation}
Thus $G_t$ can be replaced by $G_t-b(c_t)$ without changing the exact on-policy expectation. A value function is the structured version of this baseline. For the undiscounted sequence-level OPD objective, define
\begin{equation}
    V^\pi(c_t)
    =
    \E_{\student}
    \left[
    \sum_{k=t}^{T}r_k
    \mid c_t
    \right].
\end{equation}
With terminal convention $V^\pi(c_{T+1})=0$, the exact temporal residual is
\begin{equation}
    \delta_t^\pi
    =
    r_t+V^\pi(c_{t+1})-V^\pi(c_t).
\end{equation}
The value-based GAE-OPD temporal-credit hypothesis for the undiscounted OPD objective is then
\begin{equation}
    \hat A_t^{\lambda}
    =
    \sum_{l=0}^{T-t}
    \lambda^l
    \delta_{t+l}^\pi,
    \qquad
    \lambda\in[0,1].
    \label{eq:gae-exact}
\end{equation}
If $V^\pi$ is exact and samples are on-policy, Eq.~\ref{eq:gae-exact} has the same expected policy-gradient direction as the sequence-level return-to-go for any $\lambda$: future residuals beyond the first have zero conditional mean under the policy, while the first residual has expectation $Q^\pi(c_t,y_t)-V^\pi(c_t)$, where $Q^\pi(c_t,y_t)=\E_{\student}[\sum_{k=t}^{T}r_k\mid c_t,y_t]$. In this idealized case, $\lambda$ changes the variance of the estimator but not the target expectation. In practice, for $\lambda<1$, an approximate $V_\phi$ introduces bootstrapping bias through value error. In the $\lambda=1$ terminal Monte Carlo limit, the residual sum telescopes to $\sum_{k=t}^{T}r_k - V_\phi(c_t)$, so under on-policy sampling and detached rewards it remains expectation-preserving for any state-only baseline, though variance may be high. Cached old-policy rewards, repeated actor updates, missing importance correction, or trust-region approximations introduce off-policy bias.

\begin{figure*}[t]
\centering
\includegraphics[width=0.94\textwidth]{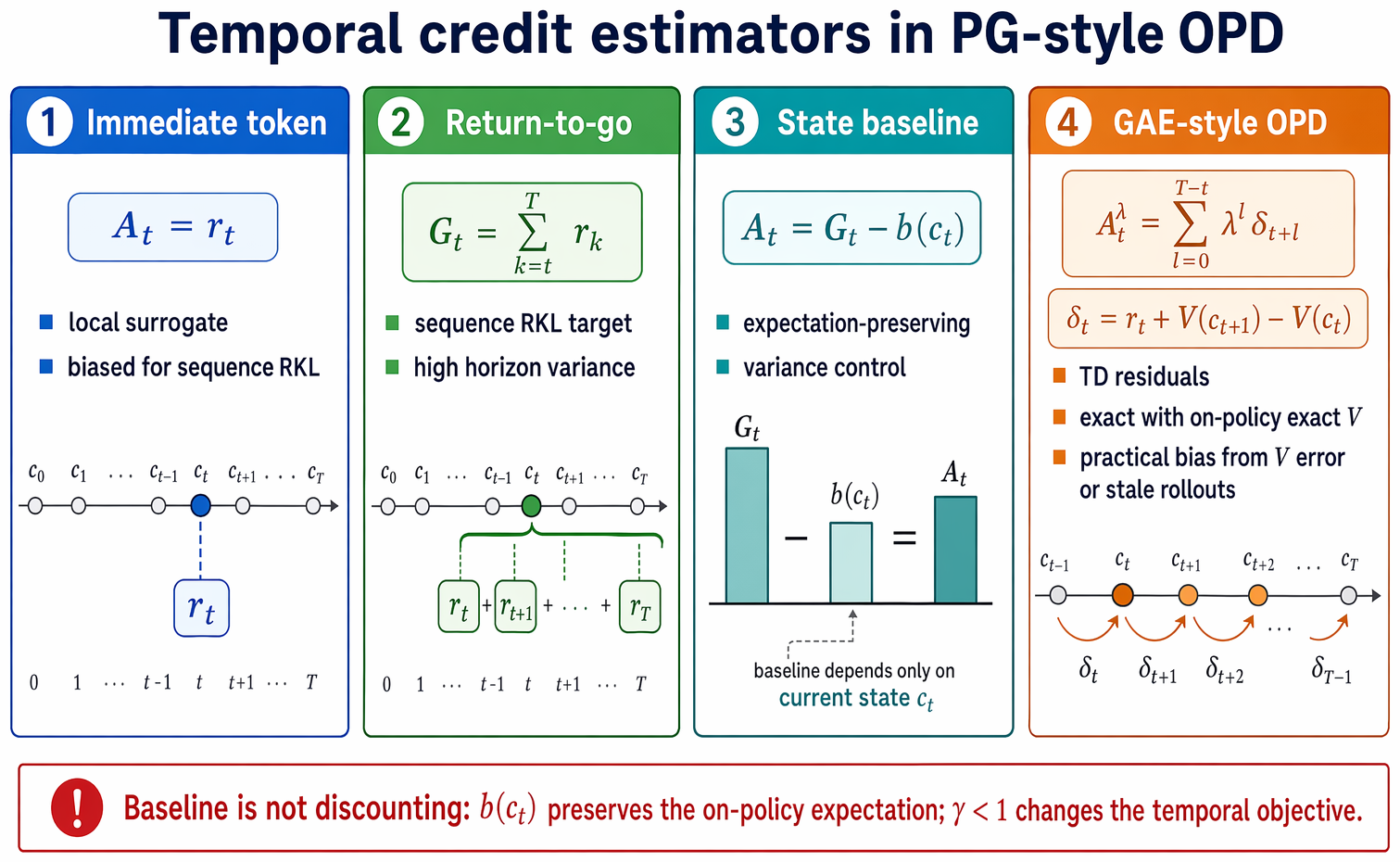}
\caption{Temporal-credit estimators for PG-style OPD. Immediate-token updates are local biased surrogates for the sequence-level reverse-KL direction; return-to-go restores suffix log-ratio credit at higher horizon variance; state-only baselines preserve the on-policy expectation; and GAE-OPD uses value-based TD residuals whose practical bias depends on critic error and policy staleness.}
\label{fig:estimator-evolution}
\end{figure*}

A rollout-buffer implementation should make this explicit. Let
\begin{equation}
    r_k^{\mathrm{old}}
    =
    \log q(y_k|c_k)
    -
    \log \oldstudent(y_k|c_k)
    \label{eq:old-reward}
\end{equation}
be the detached cached log-ratio reward computed when $\oldstudent$ generated the rollout. The critic target for the old-policy undiscounted surrogate is
\begin{equation}
    V_\phi(c_t)
    \approx
    \E_{y_{t:T}\sim \oldstudent(\cdot|c_t)}
    \left[
    \sum_{k=t}^{T}
    r_k^{\mathrm{old}}
    \mid c_t
    \right].
    \label{eq:value-target}
\end{equation}
The practical residual and advantage are
\begin{equation}
    \delta_t
    =
    r_t^{\mathrm{old}} + V_\phi(c_{t+1}) - V_\phi(c_t),
\end{equation}
and
\begin{equation}
    \hat A_t^{\mathrm{GAE}}
    =
    \sum_{l=0}^{T-t}
    \lambda^l
    \delta_{t+l}.
    \label{eq:gae-opd}
\end{equation}
The policy loss differentiates only the current student log-probability term, for example $-\sum_t \operatorname{sg}(\hat A_t^{\mathrm{GAE}})\log\student(y_t|c_t)$, optionally with PPO-style ratios against $\oldstudent$ \citep{schulman2017ppo}. Cached rewards, value targets, and advantages are treated as stop-gradient quantities in the actor update. Without small policy drift, importance correction, PPO-style ratios, or a trust-region approximation \citep{schulman2015trpo}, this old-policy surrogate should not be described as the exact current-student sequence reverse-KL gradient. If one instead defines a discounted value target
\[
    V_\gamma^\pi(c_t)
    =
    \E_{\student}
    \left[
    \sum_{k=t}^{T}\gamma^{k-t}r_k
    \mid c_t
    \right],
\]
then the corresponding residual $r_t+\gamma V_\gamma^\pi(c_{t+1})-V_\gamma^\pi(c_t)$ is unbiased for the $\gamma$-discounted surrogate when $V_\gamma^\pi$ is exact, but remains biased relative to the original undiscounted sequence-level reverse-KL objective whenever $\gamma<1$. This is why GAE-OPD is distinct from the fixed discounted estimator in Eq.~\ref{eq:discounted}: GAE uses value-based TD residuals as control variates, while discounting changes the temporal objective. \Cref{fig:estimator-evolution} summarizes these temporal-credit choices and the baseline-versus-discount distinction.

\subsection{Probability Routing under Negative Advantage}

Temporal credit is only one axis. PG-style OPD also has a vocabulary-level routing problem. If $A_t<0$, the sampled token $y_t$ is suppressed. The softmax reallocates probability mass to other tokens according to the student's current geometry, not necessarily toward teacher-supported alternatives. This is the missing counterfactual target problem.

The mechanism is visible at the logit level. For the policy loss $-A_t\log \student(a|c_t)$ on sampled token $a$, let $g_v$ denote the gradient with respect to student logit $z_v$:
\begin{equation}
    g_v
    =
    -A_t
    \left(
    \mathbf{1}[v=a]-\student(v|c_t)
    \right).
\end{equation}
Ignoring optimizer details, the first-order gradient-descent step is proportional to $-g_v$. If $A_t<0$, the sampled logit decreases, while each non-sampled logit increases in proportion to the current student probability $\student(v|c_t)$. The teacher distribution $q(v|c_t)$ affects this step only through the scalar $A_t$ unless the method explicitly adds teacher-supported alternatives. This is why suppressing a bad sampled token is not the same as routing probability mass toward a teacher-supported replacement.

Teacher top-$k$ or full-vocabulary local matching should be understood through this vocabulary-routing lens. Revisiting OPD's teacher top-$k$ support matching \citep{revisitingopd2026} keeps the update local, thereby avoiding long return-to-go variance, but its distinctive contribution is to provide a local distributional replacement target rather than a single sampled-token signal. AOPD is a related sign-conditioned response: in non-positive regions it switches from sampled-token negative reinforcement to localized divergence against teacher support \citep{aopd2026}. These support-expanded routes contribute local support and clearer probability routing, while temporal credit remains a separate estimator variable.

The gradient contrast is direct, but the normalization convention matters. For a teacher-normalized support target
\[
    q_\Omega(v|c_t)
    =
    \frac{q(v|c_t)\mathbf{1}[v\in\Omega]}
    {\sum_{u\in\Omega}q(u|c_t)}
\]
extended by zero outside $\Omega$, and for the full-vocabulary student softmax $\student$, a forward-KL or cross-entropy implementation is
\begin{equation}
\begin{aligned}
    \mathcal L_{\mathrm{FKL},\Omega}
    &=
    D_{\mathrm{KL}}(q_\Omega\|\student)
    \\
    &=
    -\sum_{v\in\Omega}q_\Omega(v|c_t)
    \log \student(v|c_t)
    +\mathrm{const},
\end{aligned}
\end{equation}
the logit gradient is
\begin{equation}
    \frac{\partial \mathcal L_{\mathrm{FKL},\Omega}}{\partial z_v}
    =
    \student(v|c_t)-q_\Omega(v|c_t),
    \label{eq:fkl-logit-gradient}
\end{equation}
Gradient descent therefore moves logits in the direction $q_\Omega(v|c_t)-\student(v|c_t)$: teacher-supported alternatives with larger target probability are explicitly increased, and overestimated alternatives are decreased. This is different from the sampled-token PG update in \cref{eq:general-pg}, where teacher information enters only through the scalar multiplying the sampled token's score term. A restricted-student variant instead defines
\[
    \pi_{\theta,\Omega}(v|c_t)
    =
    \frac{\student(v|c_t)\mathbf{1}[v\in\Omega]}
    {\sum_{u\in\Omega}\student(u|c_t)}
\]
and minimizes $-\sum_{v\in\Omega}q_\Omega(v|c_t)\log\pi_{\theta,\Omega}(v|c_t)$. For detached $\Omega$, its logit gradient is $\pi_{\theta,\Omega}(v|c_t)-q_\Omega(v|c_t)$ for $v\in\Omega$ and zero outside $\Omega$. Both teacher-target normalization and student-softmax normalization must therefore be stated.

Prior methods partially address adjacent gating and support issues by expanding support, switching objectives in non-positive regions, filtering through trust-region diagnostics \citep{tropd2026}, or repairing large teacher--student gap regimes \citep{tgpo2026}. Outcome- or trajectory-conditioned methods provide another partial precedent: SCOPE \citep{scope2026} splits correct and incorrect trajectories, SSOPD \citep{ssopd2026} turns correct--wrong rollout contrast into dense process supervision, and OPD-plus-global-feedback systems such as MiMo \citep{mimo2026} and KDRL \citep{xu2025kdrl} combine sparse correctness or reward signals with dense distillation-style guidance. Recent self-distilled RLVR and sample-selection variants further expose related variables: SD-Zero turns binary rewards into revision-derived dense supervision \citep{sdzero2026}, Self-Distilled RLVR uses self-distillation inside verifiable-reward training \citep{rlsd2026}, SRPO frames group-relative and self-distillation updates through sample routing \citep{srpo2026}, and GEAR changes the granularity of advantage reweighting for agents \citep{gear2026}. These works support sign-, outcome-, granularity-, trust-, or teacher-gap-conditioned updates; vocabulary-level replacement routing remains a separate design variable.

Our candidate routing direction is to make the replacement target explicit when the prefix is judged routable. The routing decision should be token-local even when the main OPD update uses a temporal advantage. The temporal advantage $A_t$ asks whether the sampled action contributed to downstream log-ratio returns; the local routing score below asks which replacement tokens have higher teacher-student log-ratio relative to a local baseline at the current prefix. Mixing these two roles can produce incorrect routing.

The baseline must be defined on a declared support. A full-vocabulary baseline requires full teacher scores or a smoothing convention. Under top-$k$ access, one restricted routing support is
\begin{equation}
    \Omega_t^{\mathrm{route}}
    =
    \mathrm{TopK}_T(c_t)
    \cup
    \mathrm{TopM}_S(c_t)
    \cup
    \{y_t\},
    \label{eq:routing-support}
\end{equation}
with teacher scores, student scores, and any smoothing or reconstruction convention declared on every token in $\Omega_t^{\mathrm{route}}$. Define the restricted normalized distributions
\begin{equation}
\begin{aligned}
    \tilde q_\Omega(v|c_t)
    &=
    \frac{q(v|c_t)\mathbf{1}[v\in\Omega_t^{\mathrm{route}}]}
    {\sum_{u\in\Omega_t^{\mathrm{route}}}q(u|c_t)},
    \\
    \tilde\pi_\Omega(v|c_t)
    &=
    \frac{\student(v|c_t)\mathbf{1}[v\in\Omega_t^{\mathrm{route}}]}
    {\sum_{u\in\Omega_t^{\mathrm{route}}}\student(u|c_t)} .
\end{aligned}
    \label{eq:routing-normalization}
\end{equation}
A restricted local baseline is then, writing $\Omega=\Omega_t^{\mathrm{route}}$ inside the display,
\begin{equation}
\begin{aligned}
    b_{\mathrm{loc},\Omega}(c_t)
    &=
    \sum_{u\in\Omega}
    \tilde\pi_\Omega(u|c_t)
    \ell_t^\Omega(u),
    \\
    \ell_t^\Omega(u)
    &=
    \log \tilde q_\Omega(u|c_t)
    -
    \log \tilde\pi_\Omega(u|c_t).
\end{aligned}
    \label{eq:local-routing-baseline}
\end{equation}
This is the negative reverse KL between the restricted student and teacher distributions at $c_t$; the corresponding full-vocabulary quantity can be used when full support is available. This local baseline must be distinguished from the temporal value $V^\pi(c_t)$ used for return-to-go credit. The restricted local routing score is
\begin{equation}
    \eta_t^{\mathrm{loc}}(v)
    =
    \log \tilde q_\Omega(v|c_t)
    -
    \log \tilde\pi_\Omega(v|c_t)
    -
    b_{\mathrm{loc},\Omega}(c_t).
    \label{eq:local-routing-adv}
\end{equation}
A conservative candidate set is
\begin{equation}
\begin{aligned}
    C_t^+
    =
    \{v \in \mathrm{TopK}_T(c_t)\cap \mathrm{TopM}_S(c_t):\\
    \eta_t^{\mathrm{loc}}(v)>0\}.
\end{aligned}
    \label{eq:candidates}
\end{equation}
Here $\mathrm{TopK}_T(c_t)$ denotes the $K$ highest-probability teacher tokens and $\mathrm{TopM}_S(c_t)$ denotes the $M$ highest-probability student tokens, with all candidate sets and scores treated as stop-gradient. The intersection is a reachability constraint: candidates are both teacher-endorsed and currently plausible under the student. If the intersection is empty, one may relax it to $\student(v|c_t)>\epsilon$; otherwise the update degenerates to pure suppression.

For a sampled token $u=y_t$ with $\eta_t^{\mathrm{loc}}(u)<0$, one biased corrective operator is a pairwise routing loss
\begin{equation}
\begin{aligned}
    \mathcal L_{\mathrm{route}}
    &=
    [-\eta_t^{\mathrm{loc}}(u)]_+
    \\
    &\quad\cdot
    \sum_{v\in C_t^+}
    \alpha_{t,v}
    \log(1+\exp(z_u-z_v)),
\end{aligned}
    \label{eq:routing-loss}
\end{equation}
where $z_v$ is the student logit. A concrete normalized choice is
\[
    \alpha_{t,v}
    =
    \frac{[\eta_t^{\mathrm{loc}}(v)]_+}
    {\sum_{w\in C_t^+}[\eta_t^{\mathrm{loc}}(w)]_+},
\]
when the denominator is positive, with teacher-probability or uniform weights as lower-information alternatives. If the total positive score is small, normalization can overweight marginal candidates. A temperature on the positive scores controls the concentration or peakiness of the relative routing weights; it does not by itself solve low-evidence routing. Low-evidence cases instead require a damping factor or minimum positive-mass threshold before applying the route. A temporal advantage can still gate or scale this loss, while the replacement set is defined by local teacher-student alternatives. We call this family of operators \emph{Counterfactual Routed OPD} (CR-OPD; E3). The operator is a biased corrective route by construction: it changes the local update so that negative sampled-token feedback can name teacher-supported replacements. It also requires access to teacher alternatives, so it is a support-expanded rather than logit-free method. The top-$k$ intersection is only a reachability proxy; if it is too conservative, it can miss genuinely new teacher information. Its role in this survey is to expose the probability-routing variable that standard sampled-token updates leave implicit.

\subsection{Evaluation Implications}

The decomposition above follows from the estimator algebra and yields concrete evaluation requirements. A temporal-credit study should separate four cases that are mathematically different: immediate-token OPD, full return-to-go, deliberately discounted or windowed return-to-go, and GAE-OPD with an explicit critic. It should report which estimators are unbiased for which objective, estimator variance, critic error, reward and advantage histograms, stale-policy drift, agreement with a higher-fidelity proxy when available, final task score, correct/incorrect rollout splits, length, repetition, entropy, calibration, and capability regression on golden data. If $\gamma<1$ is used, the report should state whether it is a task discount, a teacher-reliability discount, or merely a variance heuristic; only the first two define a coherent surrogate objective.

A vocabulary-routing study should hold the temporal advantage fixed while comparing plain sampled-token suppression, teacher top-$k$ direct matching, full-vocabulary matching when feasible, AOPD-style sign switching, CR-OPD-style pairwise routed alternatives, and candidate filtering. It should report support coverage, reachable-alternative rate, negative-advantage mass, teacher-student KL or log-ratio histograms, final task score, and compute overhead. Without these controls, an observed gain cannot be attributed cleanly to temporal GAE-OPD credit or to vocabulary-level probability routing.

%% file: sections/05_actionability.tex
\section{Feedback Actionability as a Diagnostic Lens}
\label{sec:actionability}

The formula taxonomy identifies the variables an OPD method can change. A separate question is whether a particular feedback signal should be converted into a gradient at a particular student-induced state. We call this \emph{feedback actionability}. Actionability is an operational diagnostic criterion: before a feedback object becomes an update, the method should justify state reliability, support reachability, credit horizon, routing target, update route, and drift control. A signal is actionable when the available evidence suggests that applying it at the visited state will improve the behavior of interest without creating unacceptable drift.

\Cref{sec:actionability,sec:failures,sec:diagnostics-recipes} use the same variables at three levels: this section defines when feedback is actionable, \cref{sec:failures} names the mechanisms that make it fail, and \cref{sec:diagnostics-recipes} turns those mechanisms into pilot diagnostics and conservative interventions.

\subsection{State Compatibility}

The first check is whether the feedback source is reliable at the student-induced state where the update is applied. This is precisely where OPD differs from teacher-forced KD: the teacher is queried on student-induced states, including mistakes. Empirical reports instantiate this risk through teacher unreliability, reasoning-pattern compatibility, marginal-information limits, token-level teachability, and tokenizer artifacts \citep{revisitingopd2026,rethinkingopd2026,manyfacesopd2026,prefixsuffix2026,rocktokens2026}; \cref{sec:failures} discusses the corresponding mechanisms.

Under our taxonomy, top-$k$ overlap, teacher perplexity on student-induced contexts, prompt-format alignment, and recoverability probes are state-compatibility diagnostics. They are risk signals whose interpretation depends on the teacher, student, support, and task. A compatible teacher can still add little marginal information; an apparently incompatible teacher can still be useful if the update route changes support, gates unreliable positions, or brings in outcome-level evidence.

\subsection{Marginal Information and Support Reachability}

Actionable feedback must contain information the student can use. Rethinking OPD and Many Faces of OPD both show that teacher--student compatibility, teacher choice, student capacity, and task domain shape whether OPD helps \citep{rethinkingopd2026,manyfacesopd2026}. If the teacher is too close to the student, OPD can degenerate into expensive self-regularization. If the teacher is too far away, the recommended tokens or reasoning pattern may be outside the student's reachable support. This is why teacher strength is not the right scalar by itself; the relevant quantity is state-conditioned marginal information under the chosen support and update route.

Support reachability makes this concrete. The support/access ladder is detailed in \cref{tab:support-cost}; the actionability question here is whether the compared alternatives are reachable under the student's state-conditioned support. Cross-tokenizer and representation work can be read as support recovery when raw teacher and student token events no longer match \citep{uld2024,dskd2025,simct2026,multilevelot2025,almcrosstokenizer2025,oprd2026}.

\subsection{Privilege Compressibility}

Some OPD variants obtain feedback from a privileged teacher: the teacher sees extra context, a verified trace, a visual crop, a domain-specialized checkpoint, an execution result, or another view unavailable to the deployed student. Privileged feedback is actionable only when the extra information changes deployable behavior rather than leaking instance-specific answers. OPCD, Vision-OPD, OPSD, GATES, and HDPO instantiate this compressibility question with context, visual, trace, consensus, or hybrid-distillation signals \citep{opcd2026,visionopd2026,opsd2026,gates2026,hdpo2026}; failure cases are separated in \cref{sec:failures}.

\subsection{Evaluation Alignment}

Dense teacher logits are local distributional feedback, not a guarantee of final task success. A token can be locally fluent while keeping the trajectory inside a wrong proof, an invalid tool plan, or a misleading answer. The evaluation-alignment check asks whether the local update agrees with the outcome, rubric, verifier, discriminator, or human criterion that actually matters.

The actionable check is whether local teacher agreement predicts the outcome criterion. Gradient-alignment, rubric, black-box, verifier, rollout-contrast, and reward-weighted variants provide probes or gates for this alignment \citep{unmaskingopd2026,ropd2026,blackboxopd2025,ovd2026,omniopd2026,mimo2026,xu2025kdrl,ssopd2026,rwopd2026,agrawal2026distributionaldagger,vpd2026}; \cref{sec:diagnostics-recipes} lists the corresponding pilot checks.

\subsection{Update Routability and Stability}

Even reliable and aligned feedback can be non-actionable if the update route is unstable. PG-style OPD can have high horizon variance; direct top-$k$ losses can introduce truncation bias; negative-advantage sampled-token updates can suppress a token without defining a replacement; self-distillation can amplify confidence or repetition \citep{certaintyopd2026,selfdistilldegrade2026,stableopd2026}.

The routability check is operationalized by the gradient-concentration, global-signal, and drift rows of \cref{tab:diagnostic-recipes}. The relevant mechanisms include variance control, non-positive-region routing, reference constraints, diagnostic gates, and calibration monitoring \citep{vopd2026,opdplus2026,aopd2026,stableopd2026,tip2026,scope2026,eopd2026,veto2026,fireopd2026,tropd2026,certaintyopd2026,selfdistilldegrade2026}.

\subsection{Evidence Boundary}

The criteria above deliberately combine direct findings with this survey's synthesis. Prior work reports sampled-token failure modes, compatibility diagnostics, length inflation, control variates, sign-based routing, and gradient-alignment diagnostics. We group those findings as state compatibility, marginal information, support reachability, privilege compressibility, evaluation alignment, and update routability. The separation between temporal credit and vocabulary routing, and the bias distinction between discounts and state-only baselines, are analytic consequences of the update formulas. GAE-OPD and CR-OPD remain hypotheses generated by those variables rather than reported implementations.

%% file: sections/06_failure_mechanisms.tex
\section{Failure Mechanisms}
\label{sec:failures}

Building on the actionability criteria in \cref{sec:actionability}, this section names the failure mechanisms that arise when a formula variable or diagnostic axis becomes unreliable. \Cref{tab:failure-mechanisms} gives the compact map; stabilization responses are deferred to \cref{sec:diagnostics-recipes}.

\begin{table*}[t]
\centering
\small
\setlength{\tabcolsep}{5pt}
\renewcommand{\arraystretch}{1.06}
\begin{tabularx}{\textwidth}{L{3.05cm}L{2.7cm}Y}
\toprule
Failure & Perturbed variable or axis & Mechanism \\
\midrule
\multicolumn{3}{@{}l}{\textit{State, feedback, and horizon}} \\
\midrule
Flawed-prefix trap & State / feedback reliability & Student enters prefixes where teacher logits are local fluency signals rather than recoverable correction. \\
\tabrowrule
Long-horizon variance & Credit estimator & Sequence-level reverse KL couples early-token updates to many noisy future log-ratios. \\
\tabrowrule
Trajectory-level support drift & State / credit under fixed interface & In multi-turn or tool-use settings, early errors move later states outside reliable teacher, verifier, or environment support. \\
\midrule
\multicolumn{3}{@{}l}{\textit{Routing and support}} \\
\midrule
Negative feedback without target & Probability routing & Suppressing a sampled token does not specify which alternative should receive the released probability mass. \\
\tabrowrule
Truncated-support bias & Support & Sampled-token or top-$k$ objectives change the full-vocabulary target and can hide important alternatives. \\
\tabrowrule
Tokenizer or modality artifact & Support / representation & Token, span, or modality events diverge semantically across teacher and student. \\
\midrule
\multicolumn{3}{@{}l}{\textit{Teacher, privilege, and system dynamics}} \\
\midrule
Cached-teacher inconsistency & Feedback source / teacher dynamics & Cached log-probabilities can become inconsistent with the current student, prompt template, tokenizer, or rollout distribution. \\
\tabrowrule
Privilege mismatch & Teacher privilege & The teacher uses context, answers, modalities, traces, or tools the student cannot internalize or access at deployment. \\
\tabrowrule
Multi-teacher conflict & Feedback source / routing & Specialist teachers provide incompatible gradients or average away weak but useful domain signals. \\
\midrule
\multicolumn{3}{@{}l}{\textit{Outcome, calibration, and regularization}} \\
\midrule
Signal-outcome mismatch & Evaluation alignment & Dense local feedback improves fluency or teacher agreement without improving the target outcome. \\
\tabrowrule
Calibration or certainty drift & Weighting / evaluation alignment & Dense feedback increases confidence or suppresses uncertainty expression while calibrated correctness does not improve. \\
\tabrowrule
Length and repetition drift & Regularizer / state & On-policy loops make repetitive continuations predictable and locally reinforced. \\
\bottomrule
\end{tabularx}
\caption{OPD failure mechanisms organized by update-interface variables and diagnostic axes. The table names mechanisms only; recipes appear in \cref{sec:diagnostics-recipes}.}
\label{tab:failure-mechanisms}
\end{table*}

\subsection{State Compatibility Failures}

The on-policy state distribution is both OPD's advantage and its risk. Student-induced states reveal what the learner must handle, but those states may also be outside the teacher's reliable correction region. Revisiting OPD reports teacher unreliability on student states and special-token or tokenizer mismatches \citep{revisitingopd2026}. Rethinking OPD reports compatibility between teacher and student thinking patterns as a key empirical factor \citep{rethinkingopd2026}. Veto reformulates targets when raw teacher guidance is unstable under a large teacher--student gap \citep{veto2026}. Recent diagnostic work on local teachability collapse and residual hard tokens points in the same direction: prefix positions, suffix positions, and particular token classes can have very different usefulness under the same nominal OPD loss \citep{prefixsuffix2026,rocktokens2026}.

\subsection{Temporal and Agentic Credit Failures}

The return-to-go estimator in \cref{eq:return-to-go} is closer to sequence-level reverse KL than immediate-token OPD, but long reasoning traces make it noisy. Immediate-token OPD drops future log-ratio terms, yielding a biased low-variance semi-gradient. vOPD, OPD+, ReOPOLD, and TrOPD intervene on different parts of this problem: variance baselines, advantage construction, reward clipping or staged exploration, and trust-region routing \citep{vopd2026,opdplus2026,reopold2026,tropd2026}. These are not interchangeable fixes; each changes a different estimator or route variable.

Agentic OPD adds discrete support drift under a fixed interaction interface. A wrong tool call, search query, or dialogue act can move the later trajectory into states where local token feedback is fluent but no longer recoverable. This mechanism concerns the update after rollouts are collected; harness fidelity is a separate interface variable. TCOD and SOD study multi-turn or step-wise OPD settings \citep{tcod2026,sodagents2026}, while GEAR and Skill-SD treat granularity or skill conditioning as explicit variables for agent self-distillation \citep{gear2026,skillsd2026}. Healthcare AI Gym provides a domain-specific reminder that agent trajectories are evaluated through interactive task processes rather than isolated next-token agreement \citep{healthcareaigym2026}. Prefix OPD, Prune-OPD, full-rollout analyses, and near-future guidance can be read as horizon-control responses to the same mechanism \citep{prefixopd2026,pruneopd2026,fullrolloutsopd2026,nearfutureopd2026}.

\subsection{Routing and Support Failures}

Negative-advantage sampled-token OPD has a vocabulary-level failure: $A_t<0$ suppresses the sampled token, but the softmax reallocates probability according to the student's current geometry rather than an explicit teacher-endorsed alternative. AOPD addresses this by switching to localized divergence in non-positive regions \citep{aopd2026}. Adjacent self-distilled RLVR, routing, and curriculum methods expose how sparse rewards, sample routes, or partial solutions can be converted into denser supervision \citep{srpo2026,sdzero2026,rlsd2026,creditopsd2026,antisd2026,paint2026,paced2026}. They motivate routed updates; the within-vocabulary replacement problem remains the specific gap analyzed in \cref{sec:credit-routing}.

Support choice is often presented as an efficiency knob, but it changes the supervision problem. The support ladder in \cref{tab:support-cost} gives the detailed regimes; the failure mechanism here is that changing support changes which alternatives are visible and how omitted mass receives gradients. DeepSeek-V4's full-vocabulary report and cross-tokenizer or representation methods illustrate the two extremes: spending systems budget for exact vocabulary support, or rebuilding support when teacher and student events are not directly comparable \citep{deepseekv42026,uld2024,dskd2025,simct2026,multilevelot2025,almcrosstokenizer2025,oprd2026}.

\subsection{Feedback Source and Evaluation Failures}

Cached-teacher systems make teacher dynamics part of the objective. Lightning OPD reduces cost by precomputing teacher log-probabilities, but its teacher-consistency condition shows why a stale teacher, prompt template, tokenizer, or rollout source changes the meaning of the stored reward \citep{lightningopd2026}. Near-policy and online variants make the same issue operational: asynchronous generation and teacher lag can improve throughput only if stale feedback remains close enough to the intended on-policy interface \citep{nearpolicy2026,oel2026}.

Privilege mismatch and signal-outcome mismatch are distinct. Privilege mismatch concerns deployability of the teacher's information: extra context, verified traces, modality views, or tool results help only when they are compressible into student behavior \citep{opcd2026,visionopd2026,opsd2026,gates2026,hdpo2026}. Signal-outcome mismatch concerns whether local teacher agreement improves the task metric; rubric, black-box, verbal, speculative, and reward-weighted variants constrain dense feedback with more semantic or outcome-oriented signals \citep{ropd2026,blackboxopd2025,ovd2026,omniopd2026,rwopd2026}. Both are actionability failures, but they require different diagnostics.

\subsection{Drift, Conflict, and Representation Artifacts}

OPD can improve the wrong observable. The Illusion of Certainty reports calibration/capability decoupling under OPD-style training, and self-distillation degradation work warns that self-generated or privileged signals can suppress useful uncertainty expression \citep{certaintyopd2026,selfdistilldegrade2026}. StableOPD identifies repetition-associated length inflation as an on-policy drift mechanism and proposes reference constraints and rollout mixtures \citep{stableopd2026}; EGRSD similarly treats self-uncertainty as a signal that should not be overwritten blindly \citep{egrsd2026}.

Multi-teacher OPD can create conflict even when each individual teacher is useful. MiMo and DeepSeek-V4 report domain-specialized teachers \citep{mimo2026,deepseekv42026}; MAD-OPD uses debate-style multi-agent feedback \citep{madopd2026}. The survey-level risk is that averaging or mixing teachers can flatten domain-specific gradients, combine incompatible reasoning styles, or hide weak signals unless teacher selection is itself routed.

Finally, tokenizer and modality artifacts are support failures in another form. Cross-tokenizer methods address text-event mismatch, while multimodal and cross-modal OPD variants such as PRISM, VOLD, CORD, VISD, Video-OPD, X-OPD, and VLA-OPD show that the state, teacher privilege, evaluation signal, or support bridge may span image, audio, video, speech, action, latent, or verifier spaces rather than a single shared vocabulary \citep{prism2026,vold2025,cord2026,visd2026,videoopd2026,xopd2026,vlaopd2026}. In these settings, the key failure is not merely noisy logits; it is an ill-defined correspondence between what the teacher evaluates and what the student can update.

%% file: sections/07_diagnostics_recipes.tex
\section{Diagnostics and Stabilization Recipes}
\label{sec:diagnostics-recipes}

The diagnostic lens is meant to change training decisions before a full OPD run fails. \Cref{tab:diagnostic-recipes} organizes recipes by formula variables rather than method names, linking each diagnostic family to warning patterns and conservative responses. The entries are pilot-study starting points; thresholds are absent because reported cutoffs are usually model-, task-, and infrastructure-specific. Unlike \cref{sec:failures}, this section is a reporting checklist for deciding what to measure or change before committing to a full run.

\begin{table*}[!t]
\centering
\footnotesize
\setlength{\tabcolsep}{4pt}
\renewcommand{\arraystretch}{1.05}
\begin{tabularx}{\textwidth}{L{2.35cm}L{3.25cm}L{4.35cm}Y}
\toprule
Diagnostic family & Probes & Warning pattern & Conservative response \\
\midrule
State and support compatibility & Top-$k$ overlap; teacher prefix perplexity; rollout-horizon probe; special-token audit. & Low or unstable overlap, high teacher perplexity on student-induced contexts, unreliable suffix positions, or special-token mismatch. & Teacher-forced SFT warmup, prompt-format alignment, smaller updates, support expansion, prefix/truncated OPD, or skipping unrecoverable positions. \\
\tabrowrule
Teacher/cache integrity & Teacher-cache consistency; prompt-template versioning; tokenizer and scoring replay. & Cached scores drift from re-scoring, teacher checkpoints or prompt templates change, or token events no longer match stored scores. & Version caches, shorten cache windows, re-score held-out slices, freeze the teacher interface, or refresh stale rollouts. \\
\tabrowrule
Gradient concentration and routability & Entropy-disagreement map; sign and mass audit; reachable-alternative check. & A few positions dominate gradients, outlier log-ratios drive updates, or negative-advantage tokens have no reachable teacher-supported alternative. & Token selection, reward clipping, advantage normalization, top-$k$ expansion, local divergence, or skipping unroutable suppression. \\
\tabrowrule
Global-signal routing & Correct/incorrect split; verifier, ORM, RLVR, rubric, discriminator, or execution-test agreement. & Dense KL improves teacher agreement while hurting solved examples, final accuracy, executable correctness, or rubric score. & Route correct and incorrect trajectories separately; gate dense OPD by outcome signals; use logit-free feedback only through an explicit route and localization rule. \\
\tabrowrule
Privilege compressibility & Held-out no-privilege evaluation; privilege ablation; deployment-context test. & Gains vanish when answer, context, visual, trace, or tool privilege is removed on held-out tasks. & Replace answer-like privilege with hints, rubrics, context summaries, verified traces, or weaker deployable signals. \\
\tabrowrule
Drift and calibration & Length, repetition, entropy, calibration, uncertainty-expression, and regression monitors. & Responses become longer, repetitive, overconfident, or less calibrated even when local distillation loss improves. & Reference KL, golden-data mixture, length control, rollout filtering, uncertainty-preserving gates, or lower OPD weight. \\
\bottomrule
\end{tabularx}
\caption{Diagnostics that convert the formula taxonomy into training decisions. The responses are conservative levers, not universal recipes.}
\label{tab:diagnostic-recipes}
\end{table*}

\subsection{A Pilot Protocol}

A conservative OPD pilot has four stages. First, collect student rollouts on held-out prompts and compute teacher log-probabilities or the available feedback signal without training. Second, run state and support diagnostics: overlap, teacher prefix perplexity, tokenizer and special-token artifacts, entropy-disagreement maps, and suffix degradation checks. Third, run a short training pilot with logging of gradient norms, log-ratio distributions, reward or advantage histograms, entropy, length, repetition, calibration, and final-task metrics. Fourth, compare correct and incorrect rollouts separately; a method that improves incorrect rollouts while damaging correct ones suggests a routing, weighting, or gate-selection problem, not necessarily a teacher problem.

An OPD report should make the following quantities auditable whenever the feedback interface permits them: final task score or verifier pass rate; teacher-student KL or sampled log-ratio histograms; support coverage and top-$k$ overlap; correct/incorrect rollout splits; prefix-versus-suffix degradation; length, repetition, entropy, calibration, and uncertainty-expression metrics; capability regression on golden data; and teacher-cache consistency or stale-score drift for cached or near-policy systems. These diagnostics help determine whether an OPD update is improving the task, merely improving teacher agreement, or moving probability mass through an unintended route.

This protocol applies to both direct-loss and PG-style OPD. For direct losses, the key decisions are support, target distribution, token weights, and regularization. For PG-style OPD, the additional decisions are advantage construction, baseline, trust-region correction, horizon control, and negative-advantage routing. For logit-free methods, the support variable is replaced by an explicit feedback route: which text span, chunk, step, action, or trajectory receives the verifier, rubric, discriminator, language-feedback, or execution signal.

\subsection{Stabilization Knobs}

Use \cref{tab:diagnostic-recipes} as the response map. State and support problems call for support expansion, horizon shortening, or skipped positions; cache problems call for versioning and re-scoring; gradient-routability problems call for clipping, baselines, advantage normalization, or local divergence; outcome and drift problems call for gates, reference constraints, golden-data mixtures, and monitoring. Method examples are summarized in \cref{tab:method-taxonomy}.

\begin{table*}[!t]
\centering
\footnotesize
\setlength{\tabcolsep}{5pt}
\setlength{\abovecaptionskip}{4pt}
\renewcommand{\arraystretch}{1.05}
\caption*{\textbf{Box 1. OPD design template.}}
\begin{tabularx}{\textwidth}{L{2.35cm}Y}
\toprule
Decision & Template entry \\
\midrule
Bottleneck & Capability transfer, expert consolidation, anti-regression, context internalization, black-box feedback, multimodal grounding, speculative drafting, long-horizon agency, or online consolidation. \\
\tabrowrule
State evidence & Test compatibility and marginal teacher information on actual student rollouts. \\
\tabrowrule
Support & If logits are available, use the smallest audited support: sampled token, teacher top-$k$, overlap support, aligned span support, or full vocabulary. \\
\tabrowrule
Feedback route & If logits are absent or brittle, localize verifier, rubric, discriminator, language feedback, speculative check, execution, representation, or multimodal signals. \\
\tabrowrule
Update route & Choose direct distributional loss, PG-style log-ratio, or a hybrid route according to support reliability and outcome-gating needs. \\
\tabrowrule
Routing diagnostics & Condition updates on correctness, entropy, disagreement, recoverability, privilege, support overlap, teacher conflict, calibration, and negative-advantage routability. \\
\tabrowrule
Stabilizers & Add reference KL, baselines, clipping, golden-data mixture, cache refresh, teacher audits, and length or calibration monitors. \\
\bottomrule
\end{tabularx}
\vspace{2pt}
\parbox{\textwidth}{\footnotesize\emph{Output:} a state-conditioned update policy specifying feedback source, support, estimator, divergence, horizon, route, and regularizer for each prefix. Fully dynamic routed OPD remains open.}
\end{table*}

%% file: sections/08_case_studies.tex
\section{Case Studies: Formula Variables in Practice}
\label{sec:case-studies}

This section uses public reports and framework documentation to show how the same formula variables appear in deployed post-training systems. Public technical reports are treated as direct-source evidence (E0), official framework documentation as implementation evidence (E1), and our variable-level readings as survey synthesis (E3). \Cref{tab:industrial-cases} names the variable or boundary illustrated by each report; detailed claim boundaries and evidence levels are audited in \cref{app:fact-audit}.

\begin{table*}[t]
\centering
\small
\setlength{\tabcolsep}{5pt}
\renewcommand{\arraystretch}{1.05}
\begin{tabularx}{\textwidth}{L{2.35cm}L{4.55cm}Y}
\toprule
System & Reported OPD role & Variable or boundary illustrated \\
\midrule
Qwen3 \citep{qwen32025} & Strong-to-weak on-policy knowledge transfer: the student generates sequences and aligns logits with larger Qwen teachers. & State source and teacher marginal information under student capacity limits. \\
\tabrowrule
MiMo-V2-Flash \citep{mimo2026} & Multi-Teacher OPD with domain-specialized teachers providing dense token-level rewards; the reported recipe can combine them with outcome-reward or GRPO-style advantages. & Domain-conditioned teacher selection and global-signal mixture: dense OPD plus ORM/RL-style advantages. \\
\tabrowrule
GLM-5 \citep{glm52026} & On-policy cross-stage distillation as final refinement; preceding-stage checkpoints serve as teachers and the stopped log-ratio replaces the GRPO advantage with group size one. & Feedback-source choice for cross-stage refinement, interpreted here as behavior preservation or anti-regression. \\
\tabrowrule
DeepSeek-V4 \citep{deepseekv42026} & Reports replacing the mixed-RL expert-merging stage with multi-teacher reverse-KL OPD over student trajectories and full-vocabulary logits. & Support fidelity and systems cost: the report uses teacher scheduling and logit-reconstruction engineering for one consolidation setting. \\
\tabrowrule
Nemotron-Cascade 2 and KAT-Coder-V2 \citep{nemotroncascade22026,katcoderv22026} & Coarser public reports describe OPD-style multi-domain or specialize-then-unify distillation inside broader cascade or agentic-coding post-training pipelines. & Breadth signal for multi-domain teacher selection and expert consolidation. \\
\bottomrule
\end{tabularx}
\caption{Industrial OPD reports viewed through the formula-driven taxonomy. Full evidence boundaries are audited in \cref{app:fact-audit}.}
\label{tab:industrial-cases}
\end{table*}

\subsection{Cross-Scale Capability Transfer}

Qwen3's technical report \citep{qwen32025} describes a strong-to-weak distillation process for smaller models using both off-policy and on-policy knowledge transfer from larger models. In the on-policy phase, the student generates sequences and is fine-tuned by aligning its logits with a teacher model; the report leaves the exact support approximation unspecified. The taxonomy highlights teacher marginal information under student capacity as the key bottleneck. A stronger teacher can provide dense guidance that a smaller model may not discover through its own exploration, while the useful part of that guidance depends on state compatibility and reachable support.

Small-model OPD makes the teacher-information variable especially sharp. Prior compatibility analyses suggest that a teacher too close to the student may add little beyond regularization, while a teacher too far from the student may point toward reasoning patterns or vocabulary choices outside the student's reachable support \citep{rethinkingopd2026,manyfacesopd2026}. The useful signal is the state-conditioned margin between what the teacher knows and what the student can absorb, not teacher strength as an isolated scalar.

\subsection{Multi-Teacher Expert Consolidation}

MiMo-V2-Flash \citep{mimo2026} and DeepSeek-V4 \citep{deepseekv42026} both report OPD in expert-consolidation settings, but they instantiate different support and systems choices. MiMo informs domain-teacher routing: it reports Multi-Teacher OPD in which domain-specialized teachers provide dense token-level rewards and the training can mix those rewards with outcome-reward advantages \citep{mimo2026}. DeepSeek-V4 informs support fidelity: it reports more than ten teacher models and full-vocabulary logit distillation intended to compute reverse-KL supervision more faithfully in that setting \citep{deepseekv42026}. Nemotron-Cascade 2 \citep{nemotroncascade22026} and KAT-Coder-V2 \citep{katcoderv22026} broaden the same theme to multi-domain or specialize-then-unify teacher selection. The shared variable is domain-conditioned teacher selection: mathematics, coding, agent, and instruction-following experts specialize in different feedback regimes. The update must decide which teacher is relevant, how much weight it receives, and how conflicts between domain behaviors are handled.

\subsection{Cross-Stage Anti-Regression}

GLM-5 \citep{glm52026} reports on-policy cross-stage distillation as a final refinement stage, using checkpoints from earlier SFT and RL stages as teachers and a stopped teacher-student log-ratio. This is a different feedback-source problem from big-to-small transfer. The teacher is not merely stronger; it is a memory of a previous training stage. In taxonomy terms, the value of the signal is behavior preservation or anti-regression, which makes reference KL, capability-specific evaluation, and prompt-format consistency especially important.

\subsection{Full-Vocabulary OPD as a Systems Case}

DeepSeek-V4 \citep{deepseekv42026} is unusually explicit about the support-fidelity problem. In Secs.~5.1.2 and 5.2.2, its report states that prior OPD work often simplifies full-vocabulary KL to token-level KL estimates reused in RL frameworks, and argues that this resource-efficient approach can lead to high-variance gradient estimates and instability. The report therefore adopts full-vocabulary logit distillation for OPD and describes systems techniques such as teacher-weight loading on demand, last-layer teacher hidden-state caching, prediction-head reconstruction of logits, teacher-index sample ordering, and specialized TileLang kernels for exact KL computation.

DeepSeek-V4 highlights support fidelity as a first-class systems variable. When the task is to consolidate many specialist teachers into one model, losing the alternative-token distribution may be more damaging than in a single-teacher, short-output setting.

\subsection{Framework Implementations as Taxonomy Checks}

As documented at the cited access dates, framework documentation shows the same variables becoming user-facing configuration knobs. SWIFT's GKD documentation \citep{swiftgkd2026} describes three sources for training responses controlled by \texttt{lmbda} and \texttt{seq\_kd}, a \texttt{beta} parameter that selects forward KL, JSD, or reverse KL, and \texttt{gkd\_logits\_topk} as a teacher-top-$k$ support approximation with renormalization; Megatron-SWIFT documentation \citep{megatronswiftgkd2026} further states that on-policy generation currently requires vLLM and otherwise falls back to off-policy responses. verl's OPD documentation \citep{verlopd2026} separates GKD OPD from PG OPD: \texttt{loss\_mode} chooses teacher-top-$k$ forward KL or sampled-token reverse-KL estimators such as k1/k2/k3, while \texttt{use\_policy\_gradient} chooses direct backpropagation or a policy-gradient advantage route. The same documentation warns against combining top-$k$ forward KL with policy-gradient aggregation; in taxonomy terms, this corresponds to the mismatch between distributional support information and sampled-token score aggregation.

The framework ecosystem is broader than SWIFT and verl. TRL's GKDTrainer documentation \citep{trlgkdtrainer2026} and DistillationTrainer documentation \citep{trldistillationtrainer2026} expose student-data fraction, sequence KD, divergence interpolation, teacher-server support, generation buffers, and top-$k$ or full-vocabulary loss controls. KDFlow \citep{kdflow2026} reports a decoupled teacher-inference and student-training system, including on-policy distillation, cross-tokenizer KD, hidden-state transfer, and student-side logit recomputation. Tinker \citep{tinkerdistillation2026} provides official cookbook examples for on-policy, multi-turn, and multi-teacher distillation. These implementation-facing facts show OPD-style variables becoming reusable infrastructure.

As operational checks, these frameworks make state or rollout source, support approximation, divergence direction, update route, teacher assignment, and rollout infrastructure concrete.

\subsection{Logit-Free, Black-Box, and Speculative Feedback}

Useful feedback can come from sources other than teacher logit vectors. Black-box OPD \citep{blackboxopd2025}, OVD \citep{ovd2026}, and rubric-based OPD \citep{ropd2026} use discriminators, verbal feedback, or task criteria when teacher internals are unavailable. Rubric-based OPD is important here because it changes the verification target: when local token probabilities miss the task criterion, structured rubrics can define what kind of behavior should be rewarded. OmniOPD \citep{omniopd2026} proposes logit-free OPD via speculative verification, moving the signal toward chunk-level or verification-derived supervision. Draft-OPD \citep{draftopd2026} applies OPD to speculative draft models by focusing on draft-induced states exposed by accepted and rejected proposals. These settings stress the feedback-source and verification variables: feedback must be constructed, calibrated, and routed even when a full teacher distribution is absent.

\subsection{Context, Multimodal, and Long-Horizon OPD}

Context distillation, multimodal self-distillation, and agentic OPD change what counts as teacher privilege. This subsection is scope-expanding evidence unless a method explicitly supplies dense feedback on student-induced states. OPCD \citep{opcd2026} uses a context-conditioned teacher to internalize useful context into a student. Vision-OPD \citep{visionopd2026} uses privileged visual views for self-distillation. Cross-tokenizer and representation-space work broadens the support question: ULD \citep{uld2024}, DSKD \citep{dskd2025}, SimCT \citep{simct2026}, MultiLevelOT \citep{multilevelot2025}, and approximate likelihood matching \citep{almcrosstokenizer2025} align token/span supports across heterogeneous tokenizers, while OPRD \citep{oprd2026} moves feedback into hidden-state space on the same rollouts. OPSD \citep{opsd2026}, self-distillation for continual learning \citep{sdft2026}, reinforcement learning via self-distillation \citep{sdpo2026}, COPD \citep{copd2026}, and OEL \citep{oel2026} raise a similar question for continual or online settings: whether the feedback source stabilizes behavior or amplifies drift. Long-horizon agentic tasks add tools, sandboxes, and delayed outcomes; those are partly interaction-interface problems, but once rollouts are collected, OPD still faces verification and credit questions. In these settings, the relevant diagnostic is whether the privilege improves grounded behavior, not merely language fluency.

%% file: sections/09_open_problems.tex
\section{Open Problems and Research Agenda}
\label{sec:open-problems}

The formula-driven view suggests open problems that are more specific than ``try another divergence.'' Each agenda item below corresponds to a variable in the feedback-to-update pipeline, including feedback allocation through temporal credit, gates or weights, and vocabulary routing.

\paragraph{Dynamic routed OPD.}
Many OPD improvements are not mutually exclusive. A mature recipe should choose support, update route, estimator, gate, teacher source, and regularizer per prefix using diagnostics such as entropy, teacher-student disagreement, advantage sign, support overlap, recoverability, and outcome correctness. The analogy is a staged RL recipe rather than a competition among global losses: OPD systems can combine support expansion, variance baselines, sign-conditioned routing, teacher selection, and drift control when their estimator assumptions are compatible. Existing methods route parts of this decision through token selection in TIP \citep{tip2026}, outcome-conditioned weighting in SCOPE \citep{scope2026}, sign-based objective switching in AOPD \citep{aopd2026}, trust-region filtering in TrOPD \citep{tropd2026}, trajectory filtering in FiRe-OPD \citep{fireopd2026}, or on/off-policy switching in AdaSwitch \citep{peng2025adaswitch}. The open problem is a coherent state-conditioned OPD policy rather than another single global loss.

\paragraph{Composability rather than a single best loss.}
The formula variables should not be read as a menu from which exactly one OPD improvement is chosen. They are analytically separable because they enter different parts of the feedback-to-update path: support determines which alternatives are visible, temporal credit determines which scalar weights a sampled action, routing determines where probability mass moves, gates decide whether the signal is trusted, and regularizers control drift. In implementation these variables are often coupled, so the constraint is not combinability in name, but compatibility of the mathematical object being optimized. A baseline can be combined with a support-expanded negative-region route; a teacher-top-$k$ loss can be combined with an outcome gate; multi-teacher routing can be combined with reference-KL stabilization. By contrast, stale rewards without correction, top-$k$ normalization without a declared support convention, or a global verifier score without a localization rule changes the estimator or feedback semantics and must be reported as such. \Cref{tab:opd-composability} expands these compatibility checks without interrupting the open-problems list.

\paragraph{Adaptive support instead of fixed support.}
Sampled-token, top-$k$, overlap, and full-vocabulary OPD are points on a support-fidelity curve. Cross-tokenizer and representation-space methods such as ULD \citep{uld2024}, DSKD \citep{dskd2025}, SimCT \citep{simct2026}, MultiLevelOT \citep{multilevelot2025}, approximate likelihood matching \citep{almcrosstokenizer2025}, and OPRD \citep{oprd2026} make the same issue explicit beyond a shared vocabulary. A useful theory should decide support per prefix: full vocabulary when alternatives matter, top-$k$ when the teacher support is concentrated, sampled-token when the signal is already stable, aligned span or byte support when tokenizers differ, and logit-free verification when logits are unavailable.

\paragraph{Counterfactual vocabulary routing.}
Negative-advantage sampled-token OPD exposes a local replacement problem that is separate from temporal credit. If a sampled token should be suppressed, the next question is which teacher-supported, student-reachable alternatives with higher local log-ratio should receive the released probability mass. Future work should define candidate-set construction, support normalization, and routing losses that use local teacher alternatives without pretending to be unbiased sequence reverse-KL estimators. Evaluations should hold the temporal advantage fixed when comparing plain suppression, teacher top-$k$ matching, full-vocabulary matching, AOPD-style switching, and pairwise routed alternatives; otherwise gains from better temporal credit and gains from better vocabulary routing are confounded.

\paragraph{OPD with outcome- or trajectory-level feedback.}
Dense teacher feedback and more global quality signals should not be mixed only by a scalar. MiMo \citep{mimo2026}, KDRL \citep{xu2025kdrl}, SCOPE \citep{scope2026}, SSOPD \citep{ssopd2026}, RW-OPD \citep{rwopd2026}, and Distributional DAgger \citep{agrawal2026distributionaldagger} provide early signals that sparse correctness, rule-based reward, reward-model scores, rubric judgments, verifier feedback, or rich expert feedback can gate or shape dense supervision. Correct trajectories, incorrect recoverable trajectories, and incorrect unrecoverable trajectories call for different updates. A rigorous OPD-plus-global-feedback recipe should specify when global feedback overrides teacher logits, when teacher logits shape local search, and when both should be ignored.

\paragraph{Credit assignment over million-token contexts.}
Long-context and agentic systems can make naive return-to-go costly and token-local OPD potentially myopic. DeepSeek-V4 \citep{deepseekv42026} exposes the systems cost of support fidelity at long context, while KETCHUP \citep{fan2025ketchup}, prefix OPD \citep{prefixopd2026}, Prune-OPD \citep{pruneopd2026}, full-rollout OPD \citep{fullrolloutsopd2026}, near-future guidance \citep{nearfutureopd2026}, TCOD \citep{tcod2026}, and SOD \citep{sodagents2026} provide early evidence that return horizon, prefix truncation, pruning, and step-wise supervision are active design variables. Future work needs lossy but faithful credit windows: which prefixes should keep raw teacher feedback, which should be summarized, and which should be evaluated only by a verifier or environment.

\paragraph{Routed OPD theory.}
Routing, gating, and support filtering intentionally introduce bias. The useful question is not whether this remains an unbiased reverse-KL estimator, but when the biased update improves task success. A minimal theory would characterize conditions under which a routed operator has lower task-level error than the nominal estimator it replaces.

\paragraph{Cross-tokenizer, multimodal, and multi-teacher OPD.}
Moving beyond a single white-box teacher with the same vocabulary makes support alignment a central problem. Cross-tokenizer OPD must decide which token or span events are comparable; multimodal OPD must decide whether the state, privilege, evaluation signal, or support bridge is textual, latent, verifier-based, or modality-specific; multi-teacher OPD must route among domain experts without averaging incompatible behaviors. Existing signals include SimCT \citep{simct2026}, MultiLevelOT \citep{multilevelot2025}, and approximate likelihood matching \citep{almcrosstokenizer2025} for cross-tokenizer KD, Vision-OPD for multimodal privilege and compressibility \citep{visionopd2026}, OPRD for representation-space feedback \citep{oprd2026}, MiMo \citep{mimo2026} and DeepSeek-V4 \citep{deepseekv42026} for explicit multi-teacher OPD, and broader cascade or agentic-coding reports such as Nemotron-Cascade 2 \citep{nemotroncascade22026} and KAT-Coder-V2 \citep{katcoderv22026} as breadth signals for specialize-then-unify consolidation.

\paragraph{Teacher recoverability estimation.}
Recoverability is easy to name and hard to measure. Top-$k$ overlap, teacher perplexity, and confidence are proxies. We need diagnostics that predict whether teacher feedback on a flawed student-induced state can still guide the student toward a valid trajectory.

\paragraph{Privilege compressibility.}
Context-, answer-, tool-, and vision-privileged teachers differ in whether their advantage can be internalized. A useful benchmark should vary privilege type while testing held-out performance without privilege, so OPD is evaluated as compression of reusable behavior rather than memorization of instance-specific hints.

\paragraph{Online OPD and slow-fast weights.}
If OPD is used for continual or online learning, full-parameter updates after every rollout may cause catastrophic drift. Continual-learning analyses make the risk concrete: sequential gradient updates can overwrite parameters important for earlier behavior unless constrained by parameter-importance penalties, replay, or restricted update subspaces \citep{kirkpatrick2017ewc}. In OPD the risk is sharper because both the visited states and the dense teacher rewards can move as the student changes. Slow-fast weight schemes \citep{ba2016fastweights}, adapters, lagged teachers, and periodic consolidation \citep{copd2026} are natural design points. Online experiential learning \citep{oel2026}, self-distillation for continual learning \citep{sdft2026}, Lightning OPD \citep{lightningopd2026}, Near-Policy \citep{nearpolicy2026}, Test-Time Speculation \citep{testtimespeculation2026}, and implementation cookbooks for on-policy or multi-turn distillation \citep{tinkerdistillation2026} provide early signals for this direction. The open question is whether OPD can be made stable enough for online deployment without relying entirely on replay memory, reference-model resets, or offline teacher-logprob caching.

\paragraph{Evaluation protocols.}
Aggregate benchmark gains hide OPD failure modes. Future evaluations should report final task score or verifier pass rate together with teacher-student KL or log-ratio histograms, support coverage and top-$k$ overlap, correct/incorrect rollout splits, prefix-versus-suffix metrics, length, repetition, entropy, calibration, uncertainty expression, capability regression on golden data, and cache consistency or stale-score drift when teacher scores are reused. These metrics should be reported by update route and support choice, because a gain from full-vocabulary direct loss, sampled-token PG, top-$k$ matching, or outcome-gated hybrid OPD can fail for different reasons.

%% file: sections/10_conclusion.tex
\section{Conclusion}
\label{sec:conclusion}

OPD is often summarized as dense teacher supervision on student rollouts. That summary is correct, but it misses why OPD is useful and sensitive to design choices: feedback is applied on student-induced states, where it may be unreliable, misaligned, high variance, poorly routed, or destabilizing. This survey complements prior method catalogs by asking how feedback becomes an auditable update.

We treated OPD as a feedback-to-update problem with allocation as a first-class part. By deriving the taxonomy from direct-loss and PG-style formulas, we made state distribution, feedback source, teacher privilege, support, temporal credit estimator, gate or weight, vocabulary probability routing, update route, and regularization explicit.

The taxonomy is generative as well as classificatory. The two directions developed here are examples of that role. First, temporal credit over student-induced trajectories is an estimator problem: discounting future log-ratios changes the temporal objective, state-only baselines preserve the on-policy expectation, and GAE-OPD is a potential TD-residual design direction rather than simple reward discounting. Second, vocabulary-level routing over local token alternatives is a support and replacement problem: negative sampled-token feedback suppresses one token without specifying where the released probability mass should go. Counterfactual Routed OPD (CR-OPD) is the corresponding potential direction on the vocabulary-routing axis.

The practical conclusion is to treat OPD as a routed feedback interface rather than a uniform loss. Its components should be chosen from diagnostics on the student states actually visited, and its future development will likely combine dynamic routing, verifier-grounded outcomes, cross-support alignment, multi-teacher selection, and online stabilization.

Future OPD reports should therefore make the checklist in \Cref{tab:reporting-checklist} explicit.

\begin{center}
\footnotesize
\setlength{\tabcolsep}{4pt}
\renewcommand{\arraystretch}{1.02}
\begin{tabularx}{\linewidth}{@{}L{2.15cm}Y@{}}
\toprule
Component & Must report \\
\midrule
State source & current policy, replay, mixed, or external \\
Feedback source & teacher, self-teacher, verifier, or outcome signal \\
Support & sampled, top-$k$, overlap, full-vocabulary, or aligned span \\
Temporal credit & immediate, return-to-go, discount, baseline, or GAE-OPD hypothesis \\
Gate/weight & correctness, entropy, disagreement, privilege, or support signal \\
Vocabulary route & sampled-token, top-$k$, switch, or CR-OPD hypothesis \\
Update route & direct loss, PG-style, hybrid, or logit-free route \\
Regularizer & reference KL, clipping, data mix, length control, or cache refresh \\
Evidence tier & paper, report, framework, or agenda \\
\bottomrule
\end{tabularx}
{\captionsetup{hypcap=false}\captionof{table}{Minimum reporting checklist for future OPD work.}}
\label{tab:reporting-checklist}
\end{center}

%% file: sections/appendix.tex
\section{Reverse-KL Score-Function Derivation}
\label{app:rkl}

For a fixed input $x$, let
\begin{equation}
    J_x(\theta)
    =
    \sum_y
    \student(y|x)
    \left[
    \log \student(y|x)-\log q(y|x)
    \right].
\end{equation}
Assume $q(y|x)>0$ whenever $\student(y|x)>0$ or that the implementation uses smoothing or truncated support. Differentiating gives
\begin{multline}
    \nabla_\theta J_x
    =
    \sum_y
    \nabla_\theta\student(y|x)
    \\
    \cdot
    \left[
    \log \student(y|x)-\log q(y|x)+1
    \right].
\end{multline}
Using $\nabla_\theta \student(y|x)=\student(y|x)\nabla_\theta\log\student(y|x)$, define
\[
a_x(y)=\log \student(y|x)-\log q(y|x)+1.
\]
Then
\begin{equation}
\begin{aligned}
    \nabla_\theta J_x
    &=
    \E_{y\sim\student}
    \Big[
    a_x(y)
    \nabla_\theta\log\student(y|x)
    \Big].
\end{aligned}
\end{equation}
The $+1$ term is a zero-mean constant score term. Removing it preserves the expected gradient under exact on-policy sampling, although it can change finite-sample variance. For gradient ascent on the negative reverse-KL cost, one may therefore use the detached reward
\begin{equation}
    R(y)
    =
    \log q(y|x)-\log\student(y|x)
\end{equation}
up to a zero-mean constant score term.

\section{Return-to-Go and Token-Local Bias}
\label{app:rtg}

When $q$ supplies autoregressive probabilities on the declared support, autoregressive factorization gives
\begin{equation}
\begin{aligned}
    R(y)
    &=
    \sum_{k=1}^{T}
    r_k,
    \\
    r_k
    &=
    \log q(y_k|c_k)-\log \student(y_k|c_k).
\end{aligned}
\end{equation}
The full sequence score is
\begin{equation}
    \nabla_\theta\log\student(y|x)
    =
    \sum_{t=1}^{T}
    \nabla_\theta\log\student(y_t|c_t).
\end{equation}
For an on-policy rollout distribution and detached rewards, past rewards can be dropped from the score term at time $t$ because they are fixed conditional on $c_t$ and the conditional score has zero mean. This yields the causal return-to-go estimator
\begin{equation}
    \hat g_{\mathrm{rtg}}
    =
    \sum_{t=1}^{T}
    \left(
    \sum_{k=t}^{T}r_k
    \right)
    \nabla_\theta\log\student(y_t|c_t).
\end{equation}
The immediate-token estimator replaces the return by $r_t$. Its expected bias relative to the sequence-level reverse-KL direction is the expectation of the dropped future term
\begin{equation}
    \E[\Delta]
    =
    \E
    \left[
    \sum_{t=1}^{T}
    \left(
    \sum_{k=t+1}^{T}r_k
    \right)
    \nabla_\theta\log\student(y_t|c_t)
    \right],
\end{equation}
where the sample-level difference is
\begin{equation}
    \Delta
    =
    \sum_{t=1}^{T}
    \left(
    \sum_{k=t+1}^{T}r_k
    \right)
    \nabla_\theta\log\student(y_t|c_t).
\end{equation}
The estimator can still be a useful semi-gradient for a local surrogate on collected prefixes, but it should not be described as an unbiased sequence-level estimator.

\section{Bias Conditions for a Proposed GAE-Style OPD Estimator}
\label{app:gae}

GAE-OPD is a value-based temporal-credit hypothesis adapted from generalized advantage estimation \citep{schulman2016gae} and distinct from simple reward discounting. For the undiscounted sequence-level reverse-KL objective, define
\begin{equation}
    V^\pi(c_t)
    =
    \E_{\student}
    \left[
    \sum_{k=t}^{T} r_k
    \mid c_t
    \right],
\end{equation}
with terminal value $V^\pi(c_{T+1})=0$. The exact TD residual is
\begin{equation}
    \delta_t^\pi
    =
    r_t+V^\pi(c_{t+1})-V^\pi(c_t),
\end{equation}
and the GAE-style advantage is
\begin{equation}
    \hat A_t^{\lambda}
    =
    \sum_{l=0}^{T-t}
    \lambda^l\delta_{t+l}^\pi .
\end{equation}
With exact $V^\pi$ and on-policy detached rewards, $\lambda$ changes variance but not the expected policy-gradient direction for the undiscounted sequence objective. The tower-property argument is
\begin{equation}
    \E[\delta_i^\pi\mid c_i]=0,
\end{equation}
and, with
\[
    Q^\pi(c_t,y_t)
    =
    \E_{\student}
    \left[
    \sum_{k=t}^{T}r_k
    \mid c_t,y_t
    \right],
\]
\begin{equation}
\begin{aligned}
    \E[
    \hat A_t^{\lambda}
    \mid c_t,y_t
    ]
    &=
    \E[
    \delta_t^\pi
    \mid c_t,y_t
    ]
    \\
    &=
    Q^\pi(c_t,y_t)-V^\pi(c_t).
\end{aligned}
\end{equation}
Future residuals vanish in conditional expectation because $\E[\delta_{t+l}^\pi\mid c_{t+l}]=0$ for $l\ge1$. Thus the expected score term is the usual advantage form for the undiscounted return-to-go. With approximate $V_\phi$, $\lambda<1$ introduces bootstrapping bias through value error, while $\lambda=1$ telescopes to Monte Carlo return minus a state baseline and does not depend on value accuracy under the terminal convention. Rewards, values, and advantages are stop-gradient quantities in the actor update unless a method explicitly defines a differentiable teacher or critic objective.

For rollout-buffer implementations, let
\begin{equation}
    r_k^{\mathrm{old}}
    =
    \log q(y_k|c_k)-\log \oldstudent(y_k|c_k)
\end{equation}
denote the detached cached log-ratio computed under the rollout policy. A practical critic may be trained toward the undiscounted cached return
\begin{equation}
    V_\phi(c_t)
    \approx
    \E_{y_{t:T}\sim\oldstudent(\cdot|c_t)}
    \left[
    \sum_{k=t}^{T} r_k^{\mathrm{old}}
    \mid c_t
    \right],
\end{equation}
with residual $r_t^{\mathrm{old}}+V_\phi(c_{t+1})-V_\phi(c_t)$ and advantage $\sum_l\lambda^l\delta_{t+l}$. This is off-policy for the current student unless the update uses importance correction, PPO-style ratios \citep{schulman2017ppo}, or a trust-region approximation \citep{schulman2015trpo}.

If one instead uses a discounted target
\begin{equation}
    V_\beta^\pi(c_t)
    =
    \E_{\student}
    \left[
    \sum_{k=t}^{T}\beta^{k-t}r_k
    \mid c_t
    \right],
\end{equation}
then $\delta_t^\beta=r_t+\beta V_\beta^\pi(c_{t+1})-V_\beta^\pi(c_t)$ and $\sum_l(\beta\lambda)^l\delta_{t+l}^\beta$ are tied to the $\beta$-discounted surrogate objective. For $\beta<1$, this surrogate is biased relative to the original undiscounted sequence-level reverse-KL objective even with an exact value function. This is the mathematical distinction between a baseline, which is expectation-preserving, and a discount, which changes future log-ratio weights.

\section{Bias Boundary for CR-OPD}
\label{app:routing}

Counterfactual Routed OPD (CR-OPD), as described in \cref{sec:credit-routing}, is a candidate vocabulary-routing operator that changes the original reverse-KL gradient in two ways. First, it restricts replacements to a candidate set $C_t^+$, often an intersection of teacher-supported and student-reachable tokens. Second, it applies a pairwise logit-ranking loss rather than the original score-function estimator. Therefore,
\begin{equation}
    \E[\nabla_\theta \mathcal L_{\mathrm{route}}]
    \neq
    \nabla_\theta
    D_t^{\mathrm{RKL}}
\end{equation}
in general as a local loss-gradient statement, and it is also a biased estimator of the sequence-level reverse-KL descent direction under the sign convention in \cref{sec:credit-routing}. CR-OPD deliberately trades unbiased reverse-KL estimation for explicit replacement-target routing: it makes explicit the otherwise implicit decision of where probability mass should go after a negative sampled-token update. Whether this biased operator improves task performance is an empirical and theoretical open problem.

\begin{table*}[!tbp]
\centering
\footnotesize
\setlength{\tabcolsep}{4pt}
\renewcommand{\arraystretch}{1.05}
\begin{tabularx}{\textwidth}{L{2.45cm}L{3.25cm}L{4.65cm}Y}
\toprule
Component & Composable dimensions & Dependency or conflict & Representative signals or methods \\
\midrule
Support fidelity & Temporal credit, gates, vocabulary routing, and regularizers. & Requires a declared support source, smoothing or normalization convention, and teacher-access budget; top-$k$ and full-vocabulary losses are not equivalent to sampled-token score estimators. & Sampled-token OPD, teacher top-$k$ matching, full-vocabulary OPD, cross-tokenizer support bridges \citep{revisitingopd2026,deepseekv42026,swiftgkd2026,uld2024}. \\
\tabrowrule
Temporal credit estimator & Support-expanded feedback, route switching, PPO-style ratios, trust regions, or reference KL. & State-only baselines preserve the on-policy expectation, while discounting, truncation, stale rewards, and approximate critics change the estimator unless the surrogate objective or correction is stated. & Immediate-token OPD, return-to-go OPD, vOPD-style control variates, proposed GAE-OPD, PPO/TRPO-style controls \citep{vopd2026,schulman2016gae,schulman2017ppo,schulman2015trpo}. \\
\tabrowrule
Vocabulary probability routing & Temporal advantages, support expansion, sign gates, and outcome gates. & Negative sampled-token feedback suppresses one token but does not name a replacement; routed alternatives require teacher-supported and student-reachable candidate sets. Routing gains should be evaluated with temporal credit held fixed. & Teacher top-$k$ local matching, AOPD non-positive-region switching, proposed CR-OPD \citep{revisitingopd2026,aopd2026}. \\
\tabrowrule
Gate or weight & Direct losses, PG-style advantages, support choice, teacher routing, and regularization. & Gates are useful only if their diagnostic signal is reliable on student-induced states; noisy gates can discard recoverable feedback or overweight misleading teacher scores. & Entropy or disagreement token selection, correctness routing, trust filtering, trajectory filtering, on/off-policy switching \citep{tip2026,scope2026,tropd2026,fireopd2026,peng2025adaswitch}. \\
\tabrowrule
Feedback source and teacher routing & Support choice, gates, outcome signals, and drift controls. & Multi-teacher or privileged feedback must handle teacher conflict, privilege compressibility, and localization of global or verifier signals before dense updates are trusted. & Multi-teacher OPD, domain-teacher routing, privileged-teacher OPD, OPD plus outcome or verifier feedback \citep{mimo2026,deepseekv42026,opcd2026,visionopd2026,xu2025kdrl}. \\
\tabrowrule
Regularizer and drift control & Direct-loss and PG-style routes, baselines, support-expanded losses, and online or cached-teacher schedules. & Over-regularization can suppress useful teacher correction, while weak regularization can cause repetition, length drift, forgetting, or stale-score drift. & Reference KL, golden-data mixture, entropy control, clipping, trust-region filtering, cache refresh \citep{stableopd2026,tropd2026,lightningopd2026,nearpolicy2026}. \\
\bottomrule
\end{tabularx}
\caption{Composability of OPD interventions under the formula-driven taxonomy. Rows are not mutually exclusive modules; they identify where an intervention enters the feedback-to-update path and which compatibility conditions must be checked when combining it with others.}
\label{tab:opd-composability}
\end{table*}
\FloatBarrier

\section{Representative Literature Matrix}
\label{app:literature-matrix}

\Cref{tab:support-cost} expands the support/access ladder summarized in \cref{sec:taxonomy}. The main text gives the compact mechanism map in \cref{tab:method-taxonomy}; \cref{tab:method-taxonomy-detailed} gives the detailed mechanism map; and the tables below provide broader literature, system, and framework coverage.

\begin{table*}[!tbp]
\centering
\footnotesize
\setlength{\tabcolsep}{4pt}
\renewcommand{\arraystretch}{1.05}
\begin{tabularx}{\textwidth}{L{2.35cm}L{3.15cm}L{3.1cm}Y}
\toprule
Support or access regime & What it saves or spends & Fidelity profile & Escalate or audit when \\
\midrule
Sampled token & Saves full-logit materialization, transfer, and storage; still needs teacher scoring or cached log-probs for sampled actions. & Lowest support fidelity; useful as a cheap PG-style signal, but blind to alternatives. & Negative updates need a replacement target, or token-local estimators are too noisy or biased \citep{revisitingopd2026,aopd2026}. \\
\tabrowrule
Teacher top-$k$ or overlap & Spends teacher top-$k$ extraction and support normalization; overlap also compares student support. & Restores local alternatives at moderate cost when teacher mass is concentrated. & Truncation hides useful alternatives, or normalization choices are not reported \citep{revisitingopd2026,swiftgkd2026,verlopd2026}. \\
\tabrowrule
Proposed student-reachable support & Spends student-side candidate filtering plus teacher scores for candidate tokens. & Intended to target alternatives the student can plausibly take for negative sampled-token feedback. & Reachability thresholds can exclude genuinely useful teacher information; empirical tradeoffs remain a research question. \\
\tabrowrule
Full vocabulary & Spends full teacher logits or systems tricks such as hidden-state caching and prediction-head reconstruction. & Highest vocabulary-level fidelity; avoids top-$k$ truncation and sampled-token replacement ambiguity. & Memory, bandwidth, or teacher scheduling dominates; DeepSeek-V4 provides one engineered feasibility point for that tradeoff \citep{deepseekv42026}. \\
\tabrowrule
Cross-tokenizer or representation support & Spends text-span, byte, lattice, optimal-transport, approximate-likelihood, or representation-space bridges. & Avoids raw-vocabulary mismatch when teacher and student token events differ. & Semantic equivalence of aligned events is assumed rather than measured \citep{uld2024,dskd2025,simct2026,multilevelot2025,almcrosstokenizer2025,oprd2026}. \\
\tabrowrule
Logit-free feedback & Saves teacher logit materialization; uses discriminator, verbal, rubric, verifier, execution, or environment feedback. & Useful under API-only or semantic-feedback settings, but no longer a raw logit KL. & The route from global or text feedback to token, span, step, or trajectory updates is unspecified \citep{blackboxopd2025,ovd2026,ropd2026,omniopd2026}. \\
\bottomrule
\end{tabularx}
\caption{Support and feedback-access choices as a systems and fidelity ladder. Higher-fidelity or better-aligned feedback generally increases materialization, transfer, alignment, or routing cost.}
\label{tab:support-cost}
\end{table*}

The support ladder hides important scaling factors beyond a simple vocabulary-size by sequence-length estimate. Long-context and multi-teacher OPD depend on teacher count, context length, batch packing, KV-cache and hidden-state reuse, prediction-head reconstruction, teacher routing, teacher freshness, and cache-refresh schedules \citep{deepseekv42026,mimo2026,lightningopd2026,nearpolicy2026}. Systems claims should therefore specify which costs are reduced: teacher forward compute, logit materialization, network transfer, storage, or stale-score refresh.

When teacher and student tokenizers differ, the support variable is no longer a shared vocabulary set. A common abstraction is to compare events in a shared text-span or byte space:
\begingroup
\small
\setlength{\jot}{1pt}
\setlength{\abovedisplayskip}{4pt}
\setlength{\belowdisplayskip}{4pt}
\begin{equation}
\begin{aligned}
    \mathcal A_T(s)
    &=
    \{u_{1:m}:\operatorname{dec}_T(u_{1:m})=s\},
    \\
    \mathcal A_S(s)
    &=
    \{v_{1:n}:\operatorname{dec}_S(v_{1:n})=s\},
    \\
    \Omega_t^{\mathrm{text}}
    &=
    \{s:\mathcal A_T(s)\neq\emptyset,\mathcal A_S(s)\neq\emptyset\},
    \\
    \bar q(s|c_t)
    &=
    \sum_{u\in\mathcal A_T(s)}
    q_T(u|c_t),
    \\
    \bar\pi_\theta(s|c_t)
    &=
    \sum_{v\in\mathcal A_S(s)}
    \student(v|c_t).
\end{aligned}
\label{eq:cross-tokenizer-bridge}
\end{equation}
\endgroup
The local loss then matches $D_{\Omega_t^{\mathrm{text}}}[\bar\pi_\theta(\cdot|c_t),\bar q(\cdot|c_t)]$ rather than comparing raw teacher and student logits. This is a conceptual bridge: multi-token text events require marginalization over tokenizations, lattice construction, optimal transport, approximate likelihood matching, or another tractable surrogate before next-token logits can be compared in a shared event space. ULD \citep{uld2024}, DSKD \citep{dskd2025}, SimCT \citep{simct2026}, optimal-transport \citep{multilevelot2025}, and approximate-likelihood-matching methods \citep{almcrosstokenizer2025} can be read as different approximations to this support bridge.

\begin{table*}[!tbp]
\centering
\scriptsize
\setlength{\tabcolsep}{4pt}
\renewcommand{\arraystretch}{1.0}
\begin{tabularx}{\textwidth}{L{2.55cm}L{2.55cm}YY}
\toprule
Work or family & Variable changed & Mechanism reading & Survey use / remaining question \\
\midrule
\multicolumn{4}{@{}l}{\textit{Foundations and diagnostics}} \\
\midrule
GKD, MiniLLM, DistiLLM/DistiLLM-2 \citep{agarwal2024onpolicy,minillm2023,distillm2024,distillm22025} & State distribution; divergence; update route. & Move supervision onto student or mixed prefixes; compare forward/reverse/JSD or reverse-KL PG-style estimators; use buffers, mixtures, or contrastive objectives for stability. & Grounds the state, objective, and route variables; teacher reliability and long-horizon routing require additional diagnostics. \\
\tabrowrule
Revisiting, Rethinking, Many Faces, Unmasking \citep{revisitingopd2026,rethinkingopd2026,manyfacesopd2026,unmaskingopd2026} & Diagnostics for state, feedback, support, and gradient alignment. & Test whether teacher feedback remains useful on student-induced contexts through sampled-token bias, teacher reliability, thinking-pattern compatibility, marginal teacher information, and per-token gradient alignment. & Grounds the actionability diagnostics; each proxy remains model-, task-, and support-dependent. \\
\midrule
\multicolumn{4}{@{}l}{\textit{Estimator and route interventions}} \\
\midrule
vOPD and OPD+ \citep{vopd2026,opdplus2026} & Credit estimator; advantage construction; stop-gradient choice. & vOPD subtracts a closed-form control-variate baseline from sampled-token OPD; OPD+ revisits how divergence-derived rewards should become advantages. & Grounds estimator and advantage variables; teacher reliability and support mismatch remain separate axes. \\
\tabrowrule
AOPD, G-OPD, TGPO, ReOPOLD, TrOPD \citep{aopd2026,gopd2026,tgpo2026,reopold2026,tropd2026} & Update route; sign-conditioned switching; reward extrapolation; teacher-guided repair; trust region. & AOPD switches non-positive regions from negative reinforcement to local divergence matching; G-OPD adds reward extrapolation beyond teacher imitation; TGPO changes the route when large teacher--student divergence makes raw log-ratio feedback poorly actionable; ReOPOLD relaxes strict imitation; TrOPD filters updates through trust-region constraints. & Grounds route interventions that can combine with baselines, support choices, and dynamic gates. \\
\midrule
\multicolumn{4}{@{}l}{\textit{Weighting and adaptive objectives}} \\
\midrule
StableOPD, TIP, SCOPE, FiRe-OPD \citep{stableopd2026,tip2026,scope2026,fireopd2026} & Weighting, filtering, and regularization. & Add reference constraints, rollout mixtures, entropy/disagreement token selection, correct/incorrect trajectory weighting, trajectory filtering, or token reweighting. & Evidence supports stability gates in specific settings; thresholds remain model- and task-dependent. \\
\tabrowrule
EOPD, ToDi, Veto \citep{eopd2026,jung2025todi,veto2026} & Divergence or target choice. & Make divergence or target reformulation conditional on local distributional geometry, teacher entropy, or instability of the raw teacher distribution. & Grounds adaptive objective selection under local distributional diagnostics. \\
\midrule
\multicolumn{4}{@{}l}{\textit{Support, systems, and feedback-source extensions}} \\
\midrule
DeepSeek-V4, SWIFT, verl, Lightning OPD, Near-Policy \citep{deepseekv42026,swiftgkd2026,verlopd2026,lightningopd2026,nearpolicy2026} & Support fidelity; teacher serving; cache freshness; rollout lag. & Move between sampled-token, top-$k$, and full-vocabulary feedback; expose teacher-serving, cache-consistency, and stale-score tradeoffs as part of the objective interface. & Makes support cost and teacher freshness auditable; exact systems choices should not be collapsed into a single OPD label. \\
\tabrowrule
Black-box OPD, OVD, ROPD, OmniOPD, KDRL, SSOPD \citep{blackboxopd2025,ovd2026,ropd2026,omniopd2026,xu2025kdrl,ssopd2026} & Feedback source; localization; outcome gate. & Replace raw teacher logits or supplement them with discriminator, rubric, verbal, verifier, outcome, or rollout-contrast feedback. & Shows when OPD becomes hybrid: global signals must state how they are routed to tokens, spans, steps, or trajectories. \\
\tabrowrule
OPCD, OPSD, Vision-OPD, ULD/DSKD/SimCT-style support bridges \citep{opcd2026,opsd2026,visionopd2026,uld2024,dskd2025,simct2026} & Privileged teacher; cross-tokenizer or representation support. & Convert extra context, answer traces, visual views, or mismatched token events into deployable student behavior through a shared support or representation bridge. & Privilege and support bridges are useful only when the extra signal is compressible into inference-time behavior and comparable under the chosen event space. \\
\bottomrule
\end{tabularx}
\caption{Detailed mechanism matrix for representative OPD, KD, and diagnostic families. The compact main-text matrix in \cref{tab:method-taxonomy} gives the reader-facing overview; this table keeps the variable-level mechanism reading auditable.}
\label{tab:method-taxonomy-detailed}
\end{table*}

\Cref{tab:lit-methods-core,tab:lit-methods-feedback-extensions,tab:lit-methods-support-extensions,tab:lit-systems-industrial,tab:lit-systems-frameworks} give compact matrices of representative work under the variables used in the main text. The matrices summarize representative coverage by mechanism and are intentionally selective rather than paper-by-paper.

\begin{table*}[!tbp]
\centering
\scriptsize
\setlength{\tabcolsep}{3pt}
\renewcommand{\arraystretch}{0.96}
\begin{tabularx}{\textwidth}{L{3.0cm}YYYY}
\toprule
Method or family & State source & Feedback and support & Objective or credit & Routing/gate and failure addressed \\
\midrule
\multicolumn{5}{@{}l}{\textit{Foundations and diagnostics}} \\
\midrule
MiniLLM \citep{minillm2023}, DistiLLM \citep{distillm2024}, DistiLLM-2 \citep{distillm22025}, GKD \citep{agarwal2024onpolicy} & Externally supplied states plus self-generated outputs in GKD-style training. & Teacher distributions or contrastive distillation signals. & Reverse-KL, contrastive, or generalized distillation objectives. & Moves distillation toward student-induced states; exposes state-distribution gap. \\
\tabrowrule
State-distribution and rich-feedback framing: Post-Training is About States \citep{nie2026states} and Distributional DAgger \citep{agrawal2026distributionaldagger} & Learner-induced states as the object of post-training analysis. & Rich expert, teacher, or distributional feedback on visited states. & Adjacent theoretical framing rather than a specific OPD optimizer. & Supports the survey's state-distribution axis and clarifies how state source differs from loss choice. \\
\tabrowrule
Revisiting OPD \citep{revisitingopd2026}, Rethinking OPD \citep{rethinkingopd2026}, Many Faces \citep{manyfacesopd2026}, Unmasking OPD \citep{unmaskingopd2026} & Student rollouts and reasoning states. & Teacher logits on student-induced contexts; support often sampled or top-$k$; gradient-alignment probes. & Token-local OPD recipes and empirical diagnostics. & Teacher compatibility, marginal information, teacher choice, loss choice, capacity sensitivity, and token-level usefulness diagnostics. \\
\midrule
\multicolumn{5}{@{}l}{\textit{Estimator, route, and horizon interventions}} \\
\midrule
AOPD \citep{aopd2026}, vOPD \citep{vopd2026}, OPD+ \citep{opdplus2026}, ReOPOLD \citep{reopold2026}, TrOPD \citep{tropd2026} & Student sampled tokens. & Teacher-student log-ratio, sometimes with local teacher support. & PG-style reward, control-variate baselines, revised advantage design, localized divergence in non-positive regions, trust-region hybrids. & Reduces variance or changes negative-advantage behavior; addresses directionless suppression and unrecoverable states. \\
\tabrowrule
Temporal-credit and horizon-control bridge: KETCHUP \citep{fan2025ketchup}, Prefix OPD \citep{prefixopd2026}, Prune-OPD \citep{pruneopd2026}, full-rollout analysis \citep{fullrolloutsopd2026}, near-future guidance \citep{nearfutureopd2026} & Sequential KD or student rollouts with varied prefix and rollout windows. & Teacher or distillation feedback over prefixes, pruned suffixes, or near-future windows. & K-step, prefix-only, pruned, truncated, or near-future credit. & Treats horizon as an operational variable; KETCHUP is a temporal-credit bridge, not core OPD. \\
\midrule
\multicolumn{5}{@{}l}{\textit{Weighting, adaptive objectives, and cautionary diagnostics}} \\
\midrule
StableOPD \citep{stableopd2026}, TIP \citep{tip2026}, SCOPE \citep{scope2026}, EOPD \citep{eopd2026}, ToDi \citep{jung2025todi}, Veto \citep{veto2026}, FiRe-OPD \citep{fireopd2026}, AdaSwitch \citep{peng2025adaswitch} & Student rollouts with stability monitoring or stage-dependent state source. & Teacher logits plus entropy, disagreement, correctness, quality, calibration, or switching signals. & Weighted, regularized, divergence-adaptive, filtered, or on/off-policy-switched OPD objectives. & Token selection, trajectory filtering, adaptive weighting, target reformulation, reference constraints, entropy-aware gates, and stage switching for noisy feedback and drift. \\
\tabrowrule
Calibration and teachability diagnostics: certainty drift \citep{certaintyopd2026}, self-distillation degradation \citep{selfdistilldegrade2026}, Prefix Teach/Suffix Fade \citep{prefixsuffix2026}, Rock Tokens \citep{rocktokens2026}, EGRSD \citep{egrsd2026} & OPD or self-distilled reasoning states. & Confidence, uncertainty, suffix teachability, residual hard-token, or self-uncertainty signals. & Diagnostic or uncertainty-aware updates. & Warns that fluent local imitation can diverge from calibrated correctness, epistemic expression, or token-level teachability. \\
\tabrowrule
Veto \citep{veto2026}, ReOPOLD \citep{reopold2026}, TrOPD \citep{tropd2026}, G-OPD \citep{gopd2026} & Student or replayed on-policy states. & Adaptive targets, relaxed objectives, trust-region constraints, or reward extrapolation. & Target reformulation and stabilization around OPD updates. & State instability, extrapolation beyond teacher, and update drift. \\
\bottomrule
\end{tabularx}
\caption{Representative core method families mapped to formula variables. Some entries are broader distillation or post-training methods included because they clarify a variable used by OPD; rows are not mutually exclusive and should not all be read as identical OPD algorithms.}
\label{tab:lit-methods-core}
\end{table*}

\begin{table*}[!tbp]
\centering
\scriptsize
\setlength{\tabcolsep}{3pt}
\renewcommand{\arraystretch}{0.96}
\begin{tabularx}{\textwidth}{L{3.0cm}YYYY}
\toprule
Method or family & State source & Feedback and support & Objective or credit & Routing/gate and failure addressed \\
\midrule
\multicolumn{5}{@{}l}{\textit{Logit-free and outcome-routed feedback}} \\
\midrule
Black-box OPD \citep{blackboxopd2025}, OVD \citep{ovd2026}, ROPD \citep{ropd2026}, OmniOPD \citep{omniopd2026} & Student rollouts with unavailable or brittle teacher logits. & Discriminator, verbal feedback, rubric, chunk-level or speculative verification. & Logit-free or criterion-level distillation. & Replaces raw logit matching with verifier, rubric, or semantic feedback. \\
\tabrowrule
Outcome-routed and self-distilled RLVR hybrids: SD-Zero \citep{sdzero2026}, Self-Distilled RLVR \citep{rlsd2026}, SRPO \citep{srpo2026}, CREDIT \citep{creditopsd2026}, AntiSD \citep{antisd2026}, PAINT \citep{paint2026}, PACED \citep{paced2026} & Self-generated reasoning samples, verifiable-reward rollouts, or competence-frontier samples. & Binary rewards, revision traces, partial solutions, sample routes, or input-specific credit; often no white-box teacher logits. & Converts sparse or global signals into denser self-distillation or routed policy updates. & Provides hybrid or adjacent evidence for outcome-conditioned dense supervision and sample routing. \\
\midrule
\multicolumn{5}{@{}l}{\textit{Draft and cached-teacher systems}} \\
\midrule
Draft-OPD \citep{draftopd2026} and Lightning OPD \citep{lightningopd2026} & Draft-induced states or offline on-policy rollouts. & Target-model feedback, accepted/rejected speculative proposals, or cached teacher scores. & OPD adapted to speculative or offline post-training workflows. & Addresses draft-state mismatch and systems cost of repeated teacher scoring. \\
\midrule
\multicolumn{5}{@{}l}{\textit{Privileged and teacher-gap feedback}} \\
\midrule
OPCD \citep{opcd2026} and Vision-OPD \citep{visionopd2026} & Student states without privileged deployment context. & Context-conditioned or vision-privileged teacher feedback. & Privileged self-distillation on shared output space. & Tests whether context or visual privilege is compressible into deployable behavior. \\
\tabrowrule
Privileged-context and hybrid-policy variants: GATES \citep{gates2026}, HDPO \citep{hdpo2026}, TGPO \citep{tgpo2026} & Student rollouts under privileged context, hybrid distillation, or large teacher--student divergence. & Consensus-gated privileged context, privileged self-distillation, or teacher-guided policy feedback. & Gates or reshapes teacher guidance when privilege or divergence makes raw OPD risky. & Broadens the teacher-privilege and teacher-gap variables and motivates explicit gates. \\
\bottomrule
\end{tabularx}
\caption{Representative feedback-source, outcome-hybrid, draft/cached-teacher, and privilege-extension families mapped to formula variables.}
\label{tab:lit-methods-feedback-extensions}
\end{table*}

\begin{table*}[!tbp]
\centering
\scriptsize
\setlength{\tabcolsep}{3pt}
\renewcommand{\arraystretch}{0.96}
\begin{tabularx}{\textwidth}{L{3.0cm}YYYY}
\toprule
Method or family & State source & Feedback and support & Objective or credit & Routing/gate and failure addressed \\
\midrule
\multicolumn{5}{@{}l}{\textit{Support-space bridges}} \\
\midrule
Cross-tokenizer and representation distillation: ULD \citep{uld2024}, DSKD \citep{dskd2025}, SimCT \citep{simct2026}, MultiLevelOT \citep{multilevelot2025}, approximate likelihood matching \citep{almcrosstokenizer2025}, and OPRD \citep{oprd2026} & Teacher and student may use different token event spaces or hidden-state feedback. & Aligned token, span, dual-space, optimal-transport, approximate-likelihood, or representation support. & Cross-tokenizer losses, recovered supervision, or representation alignment. & Prevents support artifacts where token events diverge from semantic events; expands feedback beyond output logits. \\
\midrule
\multicolumn{5}{@{}l}{\textit{Multimodal, online, and agentic settings}} \\
\midrule
Multimodal and cross-modal OPD: PRISM \citep{prism2026}, VOLD \citep{vold2025}, CORD \citep{cord2026}, VISD \citep{visd2026}, Video-OPD \citep{videoopd2026}, X-OPD \citep{xopd2026}, VLA-OPD \citep{vlaopd2026} & Student rollouts in multimodal, video, audio, speech, or action spaces. & Black-box multimodal feedback, LLM-to-VLM reasoning transfer, audio-text reasoning signals, video grounding signals, speech alignment, or VLA feedback. & Cross-modal distillation, weighted OPD, structured self-distillation, or OPD before/with RL. & Makes support alignment and grounding first-class variables; state, privilege, evaluation, or support bridges may extend beyond text vocabulary space. \\
\tabrowrule
Self/continual and co-evolving distillation: OPSD \citep{opsd2026}, SDFT \citep{sdft2026}, SDPO \citep{sdpo2026}, COPD \citep{copd2026}, and OEL \citep{oel2026} & Self-generated, experiential, or continual-learning streams. & Self-teacher, previous policy, experience-conditioned teacher, or co-evolving policy feedback. & Self-distillation, policy-distillation, or continual updates. & Raises online stability, lagged-teacher, and anti-forgetting questions. \\
\tabrowrule
Agentic granularity and online systems: GEAR \citep{gear2026}, Skill-SD \citep{skillsd2026}, Healthcare AI Gym \citep{healthcareaigym2026}, Near-Policy \citep{nearpolicy2026}, Test-Time Speculation \citep{testtimespeculation2026} & Agent, skill-conditioned, medical-agent, asynchronous near-policy, or deployment-time speculative states. & Advantage reweighting at multiple granularities, skill-conditioned self-feedback, environment or task feedback, stale near-policy rollouts, or speculative checks. & Granularity-adaptive routing, skill-conditioned self-distillation, near-policy distillation, or test-time assisted updates. & Highlights that OPD variables must be scheduled over steps, skills, and teacher freshness, especially outside single-turn text generation. \\
\midrule
\multicolumn{5}{@{}l}{\textit{Uncertainty, compression, and verifier weighting}} \\
\midrule
Uncertainty, compression, and reward-weighted variants: EGRSD \citep{egrsd2026}, CRISP \citep{crisp2026}, VPD \citep{vpd2026}, RW-OPD \citep{rwopd2026} & Self-distilled reasoning, compressed reasoning traces, language-feedback settings, or verifier-scored generation. & Self-uncertainty, iterative self-policy compression, language feedback, or property-equivalence verifier rewards. & Uncertainty-aware self-distillation, reasoning compression, variational policy distillation, or reward-weighted OPD. & Supports calibration, compression, and verifier-alignment diagnostics in the main text. \\
\bottomrule
\end{tabularx}
\caption{Representative support-alignment, multimodal, online, and diagnostic extension families mapped to formula variables.}
\label{tab:lit-methods-support-extensions}
\end{table*}

\begin{table*}[!tbp]
\centering
\footnotesize
\setlength{\tabcolsep}{4pt}
\renewcommand{\arraystretch}{1.05}
\begin{tabularx}{\textwidth}{L{3.0cm}YYYY}
\toprule
System or implementation & State source & Feedback and support & Objective or credit & Routing/gate and failure addressed \\
\midrule
Qwen3 \citep{qwen32025} & Student-generated on-policy sequences in strong-to-weak distillation. & Larger Qwen teacher logits; exact support approximation not inferred here. & Logit alignment in on-policy knowledge transfer. & Cross-scale capability transfer under student-capacity limits. \\
\tabrowrule
MiMo-V2-Flash \citep{mimo2026} & Student on-policy sequences. & Domain-specialized teachers provide dense token-level rewards. & Multi-Teacher OPD combined with outcome-reward or GRPO-style advantages. & Domain-conditioned teacher selection and outcome alignment across domains. \\
\tabrowrule
GLM-5 \citep{glm52026} & Final-stage student rollouts. & Previous-stage checkpoints as teachers. & Stopped teacher-student log-ratio used in place of GRPO advantage with group size one. & Anti-regression and behavior preservation after earlier training stages. \\
\tabrowrule
DeepSeek-V4 \citep{deepseekv42026} & Student trajectories in post-training. & More than ten domain teachers and full-vocabulary logits. & Reverse-KL multi-teacher OPD with systems support for exact KL. & Support fidelity and teacher scheduling in expert consolidation. \\
\tabrowrule
Nemotron-Cascade 2 \citep{nemotroncascade22026} and KAT-Coder-V2 \citep{katcoderv22026} & Cascade or agentic-coding post-training states. & Domain or capability-specialized teachers. & OPD-style multi-domain or specialize-then-unify distillation inside broader post-training pipelines. & Breadth signal for regression recovery, agentic capability consolidation, and domain teacher selection. \\
\bottomrule
\end{tabularx}
\caption{Representative industrial systems mapped to formula variables. The table uses public reports as evidence for usage or configuration and treats the variable-level columns as this survey's interpretation.}
\label{tab:lit-systems-industrial}
\end{table*}

\begin{table*}[!tbp]
\centering
\footnotesize
\setlength{\tabcolsep}{4pt}
\renewcommand{\arraystretch}{1.05}
\begin{tabularx}{\textwidth}{L{3.0cm}YYYY}
\toprule
System or implementation & State source & Feedback and support & Objective or credit & Routing/gate and failure addressed \\
\midrule
SWIFT \citep{swiftgkd2026} and Megatron-SWIFT \citep{megatronswiftgkd2026} & Configurable offline data, on-policy student rollouts, or teacher-generated responses. & Teacher top-$k$ logits with renormalization. & Forward KL, JSD, or reverse KL controlled by documented knobs. & Makes state or rollout source, divergence, and support approximation explicit implementation choices. \\
\tabrowrule
verl OPD \citep{verlopd2026} & GKD OPD or PG OPD workflows. & Teacher top-$k$ distributions or sampled-token reverse-KL estimators. & Direct distillation loss or PG advantage aggregation. & Separates update route from support choice; warns against incompatible combinations. \\
\tabrowrule
TRL GKDTrainer \citep{trlgkdtrainer2026}, TRL DistillationTrainer \citep{trldistillationtrainer2026}, KDFlow \citep{kdflow2026}, and Tinker \citep{tinkerdistillation2026} & Trainer or system workflows for on/off-policy, buffered, cross-tokenizer, multi-turn, or multi-teacher distillation. & Teacher servers, generation buffers, hidden-state transfer, cross-tokenizer support, or multi-teacher configuration. & Implementation-facing GKD/OPD-style trainer and system controls. & Shows OPD variables becoming reusable infrastructure across trainer and system workflows. \\
\tabrowrule
MAD-OPD \citep{madopd2026} and agentic RL settings \citep{arl2026} & Debate, tool-use, or long-horizon rollouts. & Multiple agents, tools, delayed outcomes, or task feedback. & Method-dependent post-training updates. & Highlights teacher conflict, tool/reasoning interference, harness fidelity, and long-horizon credit. \\
\bottomrule
\end{tabularx}
\caption{Representative implementation-facing systems and adjacent agentic settings mapped to formula variables.}
\label{tab:lit-systems-frameworks}
\end{table*}
\clearpage
\onecolumn

\section{Fact-Audit Levels for Claims and Design Directions}
\label{app:fact-audit}

We use four evidence levels for factual claims, scope-extension claims, and survey-derived design directions. E0 denotes direct-source evidence: an official technical report, paper, or preprint with the relevant claim. E0 records what the source reports; peer review, independent replication, and effectiveness beyond the reported setting are separate evidence questions. E1 denotes official framework or repository documentation. E2 denotes an official blog or exposition. E3 denotes this survey's interpretation, synthesis, or potential design direction. \Cref{tab:fact-audit-systems,tab:fact-audit-frameworks,tab:fact-audit-extensions} give the compact audit used for the system, framework, extension-scope, and E3 design-direction claims discussed in the main text, especially \cref{sec:case-studies}.

\begin{table}[!htbp]
\centering
\footnotesize
\setlength{\tabcolsep}{5pt}
\renewcommand{\arraystretch}{1.05}
\begin{tabularx}{\textwidth}{L{3.0cm}L{1.1cm}Y}
\toprule
Claim area & Level & Reported statement and survey use \\
\midrule
Qwen3 OPD use \citep{qwen32025} & E0 & \textbf{Reported:} strong-to-weak distillation uses both off-policy and on-policy knowledge transfer; in the OPD phase, student-generated sequences are aligned with larger Qwen teacher logits. \newline \textbf{Survey use:} evidence for cross-scale capability transfer; the exact support approximation is left unspecified by the report. \\
\tabrowrule
MiMo MOPD \citep{mimo2026} & E0 & \textbf{Reported:} Multi-Teacher OPD uses domain-specialized teachers providing dense token-level rewards, combined by default with outcome-reward or GRPO-style advantages. \newline \textbf{Survey use:} evidence for domain-conditioned teacher selection and global-signal mixture. \\
\tabrowrule
GLM-5 cross-stage OPD \citep{glm52026} & E0 & \textbf{Reported:} on-policy cross-stage distillation is used as final refinement with preceding-stage checkpoints and a stopped teacher-student log-ratio. \newline \textbf{Survey use:} evidence for checkpoint teachers and cross-stage behavior preservation as a feedback-source problem. \\
\tabrowrule
DeepSeek-V4 OPD \citep{deepseekv42026} & E0 & \textbf{Reported:} the mixed-RL expert-merging stage is replaced with multi-teacher OPD using reverse KL, student trajectories, more than ten domain teachers, full-vocabulary logit distillation, and named systems optimizations. \newline \textbf{Survey use:} evidence for support fidelity and systems cost in one expert-consolidation setting. \\
\tabrowrule
Nemotron-Cascade 2 \citep{nemotroncascade22026} and KAT-Coder-V2 \citep{katcoderv22026} & E0/E3 & \textbf{Reported:} OPD-style distillation appears inside broader cascade, multi-domain, or agentic-coding post-training systems. \newline \textbf{Survey use:} breadth evidence for multi-domain teacher selection and adjacent pipeline design. \\
\bottomrule
\end{tabularx}
\caption{Fact audit for public system claims.}
\label{tab:fact-audit-systems}
\end{table}

\begin{table}[!htbp]
\centering
\footnotesize
\setlength{\tabcolsep}{5pt}
\renewcommand{\arraystretch}{1.05}
\begin{tabularx}{\textwidth}{L{3.0cm}L{1.1cm}Y}
\toprule
Claim area & Level & Reported statement and survey use \\
\midrule
\multicolumn{3}{@{}l}{\textit{Framework and implementation evidence}} \\
\midrule
SWIFT \citep{swiftgkd2026} and Megatron-SWIFT \citep{megatronswiftgkd2026} knobs & E1 & \textbf{Reported:} official documentation exposes source-distribution, divergence, support-top-$k$, and vLLM/on-policy-generation controls. \newline \textbf{Survey use:} implementation evidence for user-facing OPD variables. \\
\tabrowrule
verl OPD knobs \citep{verlopd2026} & E1 & \textbf{Reported:} official documentation separates GKD OPD and PG OPD through \texttt{loss\_mode} and \texttt{use\_policy\_gradient}. \newline \textbf{Survey use:} implementation evidence for separating support choice from update route. \\
\tabrowrule
TRL GKDTrainer \citep{trlgkdtrainer2026}/TRL DistillationTrainer \citep{trldistillationtrainer2026}/KDFlow \citep{kdflow2026}/Tinker \citep{tinkerdistillation2026} ecosystem & E0/E1 & \textbf{Reported:} official docs and system reports expose GKD/OPD-style trainer, teacher-serving, buffering, cross-tokenizer, multi-turn, or multi-teacher workflows. \newline \textbf{Survey use:} implementation evidence for reusable OPD infrastructure. \\
\midrule
\multicolumn{3}{@{}l}{\textit{Popularization and diagnostics}} \\
\midrule
Thinking Machines OPD exposition \citep{thinkingmachines2025} & E2 & \textbf{Reported:} official exposition provides a recent public RL-compatible sampled-token OPD framing. \newline \textbf{Survey use:} context for the sampled-token PG-style route in current practice. \\
\tabrowrule
Recoverability diagnostics \citep{revisitingopd2026}, compatibility diagnostics \citep{rethinkingopd2026}, and teacher-choice diagnostics \citep{manyfacesopd2026} & E0/E3 & \textbf{Reported:} prior work reports teacher reliability, compatibility, and recoverability-related diagnostics. \newline \textbf{Survey use:} these diagnostics are organized as state-compatibility factors whose predictive value depends on the setting. \\
\bottomrule
\end{tabularx}
\caption{Fact audit for framework claims and diagnostic framing.}
\label{tab:fact-audit-frameworks}
\end{table}

\begin{table}[!htbp]
\centering
\footnotesize
\setlength{\tabcolsep}{5pt}
\renewcommand{\arraystretch}{1.05}
\begin{tabularx}{\textwidth}{L{3.0cm}L{1.1cm}Y}
\toprule
Claim area & Level & Reported statement and survey use \\
\midrule
\multicolumn{3}{@{}l}{\textit{Hybrid and extension scope}} \\
\midrule
Outcome-routed and self-distilled RLVR hybrids \citep{sdzero2026,rlsd2026,srpo2026,creditopsd2026,antisd2026,paint2026,paced2026} & E0/E3 & \textbf{Reported:} recent work converts binary rewards, partial solutions, sample routes, or input-specific credit into denser self-distillation or routed policy updates. \newline \textbf{Survey use:} evidence for outcome-conditioned gates, routes, and OPD-adjacent dense supervision. \\
\tabrowrule
Multimodal, cross-modal, and representation support \citep{prism2026,vold2025,cord2026,visd2026,videoopd2026,xopd2026,vlaopd2026,oprd2026} & E0/E3 & \textbf{Reported:} these works extend OPD-style feedback to multimodal, cross-modal, video, speech, action, or hidden-representation spaces. \newline \textbf{Survey use:} evidence that feedback space and support bridges may extend beyond text-vocabulary KL. \\
\tabrowrule
Online, near-policy, and cached-teacher variants \citep{lightningopd2026,nearpolicy2026,oel2026,testtimespeculation2026} & E0/E3 & \textbf{Reported:} these works motivate teacher freshness, cache consistency, rollout lag, and deployment-time speculation as OPD systems variables. \newline \textbf{Survey use:} evidence for treating stale feedback and cache consistency as explicit systems variables. \\
\tabrowrule
Calibration, uncertainty, and self-distillation caution \citep{certaintyopd2026,selfdistilldegrade2026,egrsd2026} & E0/E3 & \textbf{Reported:} related diagnostics warn that confidence, uncertainty expression, and calibrated correctness can move differently from local distillation loss. \newline \textbf{Survey use:} risk signals and monitoring guidance for OPD runs. \\
\midrule
\multicolumn{3}{@{}l}{\textit{Analytic distinctions and survey-derived design directions}} \\
\midrule
GAE-OPD temporal-credit hypothesis, adapted from GAE \citep{schulman2016gae} & E3 & \textbf{Analysis:} value-based TD residuals and direct discounting are mathematically different routes over teacher-student log-ratio returns; exact on-policy values preserve the undiscounted sequence objective under the stated conditions. \newline \textbf{Survey use:} GAE-OPD is proposed as a temporal-credit hypothesis; discounting future log-ratios defines a biased surrogate unless explicitly treated as the target objective. \\
\tabrowrule
Counterfactual Routed OPD & E3 & \textbf{Analysis:} CR-OPD is formulated as a biased routing operator for negative-advantage tokens. \newline \textbf{Survey use:} it makes the replacement target explicit and leaves empirical value as an open research question. \\
\bottomrule
\end{tabularx}
\caption{Fact audit for extension-scope claims and survey-derived design directions.}
\label{tab:fact-audit-extensions}
\end{table}